\documentclass[journal]{IEEEtran}
\usepackage{amsmath,amsfonts,amssymb}
\usepackage{mathtools}
\usepackage{algorithmic}
\usepackage{algorithm}
\usepackage{array}
\usepackage[caption=false,font=normalsize,labelfont=sf,textfont=sf]{subfig}
\usepackage{textcomp}
\usepackage{stfloats}
\usepackage{url}
\usepackage{verbatim}
\usepackage{graphicx}
\usepackage[acronym]{glossaries}
\usepackage[table]{xcolor}
\usepackage{hyperref}
\usepackage{microtype}
\usepackage{enumitem} 
\usepackage{booktabs}
\usepackage{multirow}
\usepackage{siunitx}
\usepackage{flushend}

\newcommand{\method}{UNSEEN}
\newcommand{\methodslam}{\method{}-SLAM}
\newcommand{\methodplan}{\method{}-PLAN}
\newcommand{\methodmap}{\method{}-MAP}
\newcommand{\methodtrajopt}{\method{}-OPT}
\newcommand{\vslam}{V-\gls{slam}}

\newcommand{\problemname}{\emph{Exploratory Point-to-Point}}
\newcommand{\freespace}{\mathcal{F}}
\newcommand{\estfreespace}{\hat{\mathcal{F}}}
\newcommand{\unknownspace}{\Omega}
\newcommand{\traj}{\Theta}
\newcommand{\estTraj}{\hat{\traj}}
\newcommand{\optTraj}{\hat{\traj}^\star}
\newcommand{\estPose}{\hat{\pose}}
\newcommand{\SEthree}{\mathrm{SE(3)}}
\newcommand{\Spheretwo}{\mathbb{S}^2}
\newcommand{\SOthree}{\mathrm{SO(3)}}
\newcommand{\SOthreealgebra}{\mathfrak{so}(3)}
\newcommand{\Rthree}{\mathbb{R}^3}
\newcommand{\RPthree}{\mathbb{RP}^3}

\newcommand{\ctrlset}{\mathcal{U}}

\newcommand{\pose}{\mathbf{T}}
\newcommand{\poset}{\pose(t)}
\newcommand{\posenew}{\pose_\text{new}}
\newcommand{\estposenew}{\estPose_\text{new}}
\newcommand{\estposerand}{\estPose_\text{rand}}
\newcommand{\estposenear}{\estPose_\text{near}}

\newcommand{\posedesfinal}{\pose_F^d}
\newcommand{\estposedesfinal}{\estPose_F^d}
\newcommand{\trajectory}{\Theta}

\newcommand{\translation}{\mathbf{t}}
\newcommand{\rotation}{\mathbf{R}}
\newcommand{\angleaxisvector}{\xi}
\newcommand{\limitangleaxisvector}{\Bar{\angleaxisvector}}
\newcommand{\figref}[1]{Fig.~\ref{#1}}

\newcommand{\slamstate}{\mathcal{X}}

\newcommand{\cameraimage}{\mathcal{I}}
\newcommand{\globalmap}{\mathcal{G}}
\newcommand{\localmap}{\mathcal{M}}

\newcommand{\keyframes}{\mathcal{K}}

\newcommand{\landmark}{\ell}

\newcommand{\projection}{\pi}

\newcommand{\covariancemat}{\Sigma}
\newcommand{\posecovariance}{\covariancemat_{\pose}}
\newcommand{\landmarkcovariance}{\covariancemat_{\landmark}}
\newcommand{\blockcovariance}{\boldsymbol{\covariancemat}}
\newcommand{\statecovariance}{\blockcovariance_{\slamstate}}
\newcommand{\camerameasurementcovariance}{\covariancemat_{\camerameasurement}}
\newcommand{\measurementcovariance}{\boldsymbol{\covariancemat}_{\stackedresidual}}
\newcommand{\poseposeblockcovariance}{\blockcovariance_{\keyframes \keyframes}}
\newcommand{\poselandmarkblockcovariance}{\blockcovariance_{\keyframes \globalmap}}
\newcommand{\landmarklandmarkblockcovariance}{\blockcovariance_{\globalmap \globalmap}}

\newcommand{\informationmat}{\Lambda}
\newcommand{\blockinformationmat}{\boldsymbol{\Lambda}}
\newcommand{\poseposeblockinformationmat}{\blockinformationmat_{\keyframes \keyframes}}
\newcommand{\poselandmarkblockinformationmat}{\blockinformationmat_{\keyframes \globalmap}}
\newcommand{\landmarklandmarkblockinformationmat}{\blockinformationmat_{\globalmap \globalmap}}
\newcommand{\stateinformationmat}{\blockinformationmat_{\slamstate}}

\newcommand{\camerameasurement}{\mathcal{C}}

\newcommand{\schurcomplement}{\mathbf{S}}

\newcommand{\jacobian}{\mathbf{J}}
\newcommand{\leftjacobian}{J_{\mathit{l}}}
\newcommand{\limitleftjacobian}{\Bar{J}_{\mathit{l}}}

\newcommand{\residual}{r}
\newcommand{\cameraresidual}{\residual_{\camerameasurement}}
\newcommand{\poseregularization}{\residual_{\pose}}
\newcommand{\landmarkregularization}{\residual_{\landmark}}
\newcommand{\stackedresidual}{\mathbf{\residual}}

\newcommand{\obstacle}{\mathcal{O}}
\newcommand{\estobstacle}{\hat{\mathcal{O}}}
\newcommand{\leafnode}{n}
\newcommand{\volume}{V}
\newcommand{\entityvolume}[1]{\volume_{#1}}

\newcommand{\visiblevolume}{\entityvolume{\textup{FOV}}}

\newcommand{\chithreshold}{\chi_{\textup{Th}}}
\newcommand{\spacevector}{x}

\newcommand{\tree}{\mathcal{T}}
\newcommand{\treeedge}{\mathcal{E}}
\newcommand{\treenode}{\nu}
\newcommand{\treepath}{\Pi}
\newcommand{\globaltreepath}{\Pi_{\textup{G}}}
\newcommand{\localtreepath}{\Pi_{\textup{L}}}
\newcommand{\nodecost}{\zeta}

\newcommand{\angularvelocity}{\omega}
\newcommand{\linearvelocity}{v}
\newcommand{\angularacceleration}{\alpha}
\newcommand{\linearacceleration}{a}
\newcommand{\constraintmatrix}{\mathbf{A}}
\newcommand{\constraintvector}{\mathbf{b}}
\newcommand{\motionconstraintmatrix}{\constraintmatrix_{\textup{M}}}
\newcommand{\motionconstraintvector}{\constraintvector_{\textup{M}}}
\newcommand{\freespaceconstraintmatrix}{\constraintmatrix_{\freespace}}
\newcommand{\freespaceconstraintvector}{\constraintvector_{\freespace}}
\newcommand{\optimizationcost}{\Gamma}

\newcommand{\polytope}{\mathcal{P}}
\newcommand{\edgepolytope}{\polytope^\treeedge}

\newcommand{\orthobasis}[2]{\mathrm{e}_{#2}^{#1}}

\newcommand{\translationctrlpoint}{\prescript{\translation}{}{\mathbf{c}}}
\newcommand{\rotationctrlpoint}{\prescript{\angleaxisvector}{}{\mathbf{c}}}

\newcommand{\numbercontrolpointstranslation}{\prescript{\translation}{}{\mathrm{K}}}
\newcommand{\numbercontrolpointsrotation}{\prescript{\angleaxisvector}{}{\mathrm{K}}}
\newcommand{\timeDep}{(t)}
\newcommand{\basisfunction}{B}
\newcommand{\controlpointsvector}{\boldsymbol{\tau}}

\newcommand{\controlpointsgradientmatrix}{\mathbf{B}}
\newcommand{\controlpointsgradientmatrixtranslation}{\prescript{\translation}{}{\controlpointsgradientmatrix}}
\newcommand{\controlpointsgradientmatrixrotation}{\prescript{\angleaxisvector}{}{\controlpointsgradientmatrix}}

\newcommand{\thrust}{\mathbf{f}}

\newcommand{\cameradirection}{d}

\newcommand{\activemap}{\localmap_{\textup{A}}}
\newcommand{\inactivemap}{\localmap_{\textup{I}}}
\newcommand{\visiblemap}{\localmap_\textup{V}}

\newcommand{\transpose}{^\mathsf{T}}
\newcommand{\dt}{^\prime}
\newcommand{\ddt}{^{\prime\prime}}
\newcommand{\wrt}{with respect to}

\newcommand{\cellwin}[1]{\cellcolor{mygreen} \textbf{#1}}
\newcommand{\celldraw}[1]{\cellcolor{myyellow} \textbf{#1}}

\newacronym{slam}{SLAM}{Simultaneous Localization and Mapping}
\newacronym{fov}{FoV}{Field of View}
\newacronym{unseen}{UNSEEN}{Uncertainty-aware Navigation via Sparse Estimation in unknown ENvironments}
\newacronym{esdf}{ESDF}{Euclidean Signed Distance Function}
\newacronym{rhp}{RHP}{Receding Horizon Planning}
\newacronym{pag}{PAG}{Perception-Agnostic}
\newacronym{paw}{PAW}{Perception-Aware}
\newacronym{MAP}{MAP}{Maximum a Posteriori}
\newacronym{MLE}{MLE}{Maximum Likelihood Estimation}
\newacronym{ekf}{EKF}{Extended Kalman Filter}
\newacronym{nbv}{NBV}{Next Best View}
\newacronym{tsdf}{TSDF}{Truncated Signed Distance Field}
\newacronym{mav}{MAV}{Micro Aerial Vehicle}
\newacronym{uav}{UAV}{Unmanned Aerial Vehicle}

\newacronym{bch}{BCH}{Baker-Campbell-Hausdorff}

\newacronym{APE}{APE}{Absolute Pose Error}
\newacronym{ATE}{ATE}{Absolute Translation Error}
\newacronym{ARE}{ARE}{Absolute Rotation Error}
\newacronym{RPE}{RPE}{Relative Pose Error}
\newacronym{RTE}{RTE}{Relative Translation Error}
\newacronym{RRE}{RRE}{Relative Rotation Error}

\newacronym{rmse}{RMSE}{Root Mean Square Error}
\newacronym{cpu}{CPU}{Central Processing Unit}
\newacronym{gpu}{GPU}{Graphics Processing Unit}

\definecolor{mygreen}{RGB}{200,255,200}
\definecolor{myyellow}{RGB}{255,255,200}

\newcommand{\eg}{e.g.,~}
\newcommand{\ie}{i.e.,~}
\newcommand{\etal}{\textit{et al.}}
\newcommand{\etalcite}[2]{{#1}~\etal~\cite{#2}}

\usepackage{pifont}
\newcommand{\cmark}{\ding{51}}%
\newcommand{\xmark}{\ding{55}}%

\begin{document}

\title{\method{}: Uncertainty-aware Navigation via Sparse Estimation in Unknown Environments
\thanks{All authors are with the University of Trento, Italy.
Email: $\{$tommaso.faraci, marco.camurri, daniele.fontanelli, luigi.palopoli$\}$@unitn.it.\\
\textsuperscript{1} Tommaso Faraci is a Ph.D. student.
}
}

\author{\IEEEauthorblockN{1\textsuperscript{st} Tommaso Faraci\textsuperscript{1}}, 
\and
\IEEEauthorblockN{2\textsuperscript{nd} Marco Camurri}, 
\and
\IEEEauthorblockN{3\textsuperscript{rd} Daniele Fontanelli},
\and
\IEEEauthorblockN{4\textsuperscript{th} Luigi Palopoli} 
}

\maketitle

\begin{abstract}
Visual navigation in unknown environments remains a core challenge in mobile robotics, especially for resource-constrained platforms. 
Most existing approaches rely on loosely coupled modular pipelines and strong assumptions on perception quality or environmental structure, often resorting to multi-modal sensor suites that increase system complexity and deployment cost. 
Vision-only navigation offers a lightweight alternative, but its performance degrades severely under motion blur, low texture, and illumination changes, largely because they neglect the tight coupling between commanded motion and perception. 
While perception-aware methods partially address this issue, they typically optimize individual modules and fail to propagate uncertainty consistently across the navigation stack. 
In this paper, we present \method{}, a unified uncertainty- and perception-aware navigation framework that explicitly couples localization, mapping, and planning using only a front-mounted camera. 
\method{} estimates sparse maps and robot poses with associated uncertainties at 6~Hz, and leverages them to plan trajectories that jointly optimize task progress and estimation accuracy in receding-horizon.
Simulations and extensive real-world experiments in unknown environments demonstrate the robustness of the proposed approach, with \methodslam{} reducing absolute translational error by 9.8\% and \methodplan{} improving estimation accuracy by up to 45\% compared to state-of-the-art methods, while achieving a 100\% task success rate.

\end{abstract}
\section{Introduction}
\label{sec:intro}

\IEEEPARstart{A}{utonomous} navigation in unknown environments remains a fundamental challenge in robotics for resource-constrained \glspl{uav}. In the absence of prior knowledge, perception, mapping, and planning are tightly coupled, such that even minor estimation errors can rapidly compromise safety and mission success.

Perception-aware navigation methods partially address this coupling, but typically optimize isolated components of the autonomy stack. To maintain tractability, many approaches rely on strong assumptions, including prior maps~\cite{tordesillas_panther_2022, bartolomei_semantic-aware_2021}, near-perfect localization~\cite{zhou2021fuel, tordesillas_panther_2022, kondo2024puma}, or noise-agnostic models that neglect estimation uncertainty~\cite{zhang_falcon_2024, kondo_dynus_2025}. Other solutions mitigate drift through multimodal sensing~\cite{spechtLocalizationAware, chen_apace_2024}, at the cost of increased hardware complexity and computational overhead, limiting their applicability to lightweight aerial platforms.

Vision-only navigation offers a compelling alternative due to its low power consumption and minimal sensor footprint~\cite{loquercio2021learning, kaufmann_champion_level_2023}. However, most visual pipelines remain brittle under motion blur, low texture, or lighting variations. As a result, visual navigation is often integrated with other sensors and relies on dense environment representations, promoting decoupled estimation and planning pipelines that partially or entirely neglect the probabilistic nature of state estimation. 

In this work, we present \method{}, a vision-only framework for navigation in unknown environments that explicitly relies on sparse map representations and consistent end-to-end uncertainty propagation. The framework is built around a lightweight Visual-\gls{slam} module, \methodslam{}, which estimates sparse, uncertain maps and poses that are structured and managed within \methodmap{}. Leveraging this sparse representation, \methodplan{} and \methodtrajopt{} generate safe motion plans in free and known space, explicitly improving future estimation quality.

\begin{figure}[t]
\centering
\includegraphics[width=.9\columnwidth]{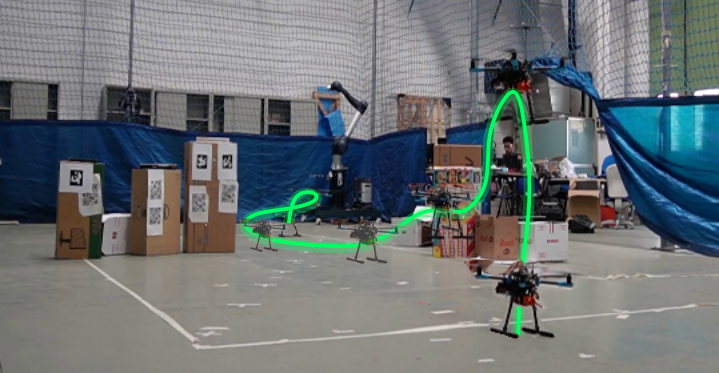}
\caption{\method{} executing a trajectory in an unknown environment containing regions of varying texture and perceptual difficulty.}
\label{fig:intro_flight}
\end{figure}

The main contributions of this work are summarized as follows:
\begin{enumerate}
\item{} A novel \vslam{} extracting state covariances from a pose graph and propagating them both up- and downstream to concurrently improve estimation conditioning and safety in mapping and planning (Sec.~\ref{sec:slam&mapping}).

\item{} A novel mapping framework for probabilistic sparse maps employing truncated Gaussian distributions, enabling occupancy updates and uncertainty-aware planning (Sec.~\ref{sec:slam&mapping}).

\item{} An uncertainty-aware receding horizon planning algorithm for sparse visual navigation in unknown environments that ensures safety and estimation robustness, combining global exploration planning for exploration with local smoothing (Sec.~\ref{sec:planning}).

\item{} A novel constrained \gls{MLE} problem on $\SEthree$ for trajectory optimization maximizing the effectiveness of visual navigation under uncertainty with a novel visibility model (Sec.~\ref{sec:trajopt}).

\item{} A tightly-integrated and unified autonomy stack that ensures vision-only safe and reliable navigation for resource-constrained platforms.
\end{enumerate}

\noindent To the best of our knowledge, \method{} is the first framework to achieve online, uncertainty-aware, perception-aware planning and trajectory optimization in unknown environments using only a front facing camera for sparse estimation.

\section{Related Work}
\label{sec:related_work}
Literature on autonomous visual navigation in unknown environments is vast, although it can be categorized based on the single component of the pipeline each contribution mostly focus on: Localization and Mapping, Planning and Trajectory Optimization. Hence, we report the main contributions in each category, providing a summary of the most relevant works compared to \method{} in Table~\ref{tab:algos}.

\subsection{SLAM and Map representations for Navigation}
\label{sec:related_work:slam}
\label{sec:related_work:mapping}

Modern autonomous navigation pipelines rely on visual state estimation
to infer both robot motion and a representation of the surrounding
environment.
\vslam{} systems jointly estimate camera poses and scene structure
from image measurements leveraging probabilistic inference.
In general, \vslam{} backends can be broadly
categorized into filtering- and smoothing-based.

\subsubsection{Sparse Visual and Visual-Inertial SLAM}
Filtering-based methods estimate the state recursively through Bayesian
updates, which requires to maintain state covariances over time to perform predictions and corrections.
MonoSLAM~\cite{davison2007monoslam}, ROVIO~\cite{bloesch2017rovio}, OpenVINS~\cite{Geneva2020openvins} and other proposed approaches~\cite{mourikis2007multi} provide principled uncertainty propagation, but scale
poorly with map size, limiting long-term or
large-scale deployments.

In contrast, smoothing-based methods propose factor-graph optimization over a selected set of keyframes and landmarks in a nonlinear least-squares framework.
For example, Kimera~\cite{rosinol2020kimera}, OKVIS2-X~\cite{okvis2} and VINS-Mono~\cite{qin2019b} employ smoothing backends such as g2o~\cite{grisetti2011g2o}, iSAM2~\cite{isam2} or GTSAM~\cite{dellaert2017factor} to perform fixed-lag smoothing or full-batch optimization.
These methods leverage sliding-windows and factor graphs to either marginalize old poses and landmarks to limit computational complexity, or efficiently exploit sparsity during inference. Due to the re-linearization of past poses, these methods  achieve superior accuracy and scalability compared to filtering approaches \cite{okvis2,rosinol2020kimera}. 
As such, most of these systems operate on sparse feature maps but strictly rely on correcting IMU measurements for accurate localization rather than effective environment representation. 
For this reason, marginalized states are discarded after optimization.
As a result, the optimized map is rarely refined beyond its immediate
use for pose estimation.
ORB-SLAM~\cite{mur2015orb} and its successors~\cite{mur2017orb,orb_slam3} partially mitigate this limitation by
maintaining a global map atlas and reoptimizing landmarks during bundle
adjustment, but uncertainty information is still discarded after optimization.
\subsubsection{Dense \vslam{}}
Although less common in navigation, dense methods have been developed in parallel to sparse ones.
Instead of optimizing poses and landmarks, dense and semi-dense \vslam{} methods optimize dense maps, thus delivering richer environment representations.
DTAM~\cite{newcombe2011dtam} and other approaches~\cite{engel2015large} estimate camera motion by minimizing photometric
error over full images, enabling denser reconstructions at the
cost of robustness and computational load.
The introduction of RGB-D sensing made
real-time dense reconstruction possible with volumetric and surfel-based dense
mapping systems such as KinectFusion~\cite{newcombe2011kinectfusion} and ElasticFusion~\cite{whelan2016elasticfusion}.
However, most of these systems alternate between tracking within a static map and
mapping with fixed poses, which degrades trajectory accuracy at scale
and prevents their use for navigation.
Consequently, current navigation methods often decouple state
estimation and mapping, ensuring accurate pose estimates from feature-based \vslam{} and using them to update dense maps using point clouds from RGB-D, stereo cameras or lidars~\cite{han2019fiesta,hornung2013Octomap,papatheodorou2025efficientsubmapbasedautonomousmav}.

\subsubsection{Uncertainty Handling in \vslam{}}
Both sparse and dense \vslam{} systems share limitations in
the handling of uncertainty.
Most smoothing-based formulations assume uncertainty to flow
unidirectionally from sensor noise into the backend optimization, after which
state covariances are not computed or retained.
Reoptimizing landmarks without retaining their prior confidence ignores
the evolution of estimation uncertainty and prevents its exploitation
by downstream modules such as mapping or planning.

Recent learning-based approaches have attempted to infer uncertainty
models jointly with state estimation.
DROID-SLAM \cite{teed2021droid} introduced a differentiable bundle adjustment layer to learn
feature confidence, but it relies on simplified diagonal covariance models
and focuses on relative uncertainty rather than metric consistency.

\etalcite{Qiu}{qiu2025macvo} proposed MAC-VO, which propagates learned 2D uncertainties into 3D covariances using
empirical projection models and first-order corrections, capturing complex image-level uncertainty but only
approximating its evolution through bundle adjustment.
Overall, uncertainty is either ignored, heuristically approximated, or
not preserved in a form suitable for downstream decision-making.

\subsubsection{Mapping for Navigation}
For navigation, sensor data is typically converted into an
occupancy representation to enable collision checking and path
planning.
Octomap~\cite{hornung2013Octomap} enables compact storage and multi-resolution
queries, making it a de facto standard for mobile robot
navigation.
However, it assumes deterministic and accurate pose estimates, not allowing for reoptimization of map sections from older poses.
Additionally, it only updates voxels intersected by sensor rays, neglecting the spatial
uncertainty induced by state estimation.

In practice, however, \vslam{} delivers probabilistic 3D landmark distributions. 
When subject to drift, noisy measurements and degraded visual conditions, the uncertainty spreads spatially becoming relevant to neighboring voxels.
\etalcite{Duong}{duong_autonomous_2021} proposed a kernel-based representation to model free and
occupied space continuously as distributions while accounting for sensor noise, but at the expense of scalability and
compatibility with standard planning frameworks.
\etalcite{Pairet}{pairet2022safenav} proposed a method that incorporates collision risk
into planning to account for the probabilistic nature of estimated obstacle positions, but it relies on externally provided maps and ignores state
estimation uncertainty.

Modern navigation systems integrate occupancy maps with volumetric representations such as
\glspl{tsdf}~\cite{oleynikova2017voxblox,han2019fiesta} or
hierarchical
octrees~\cite{hornung2013Octomap,vespa2018efficient,duberg2020ufomap,reijgwart2023wavemap}. These are combined with search-based~\cite{fastautonomousflight2018},
sampling-based~\cite{karaman2011sampling}, or optimization-based~\cite{zucker2013chomp,oleynikova2020open} planners.

While robust, these pipelines often execute estimation, mapping, and
planning sequentially, introducing latency and limiting reactivity by locking each process based on the outcome of the previous.
Hierarchical mapping approaches such as WAVEMAP~\cite{reijgwart2023wavemap} improve efficiency by
adapting map resolution to planning needs, but still assume deterministic state
estimates and do not incorporate estimation uncertainty.

Accounting for estimation uncertainty requires to factor estimated maps and poses and tightly couple estimation with occupancy mapping.
As stated above, while sparse \vslam{} presents a valid option for navigation tasks. However, most of the aforementioned marginalize landmarks, neglecting maps.
Attempts to merge sparse \vslam{} maps into useful environment representations have been made.
\etalcite{Chen and Liu}{chen2021navigablespace} proposed a method to convert sparse point clouds into navigable space, focusing on local free-space extraction and convex decomposition, not considering uncertainty and therefore without producing an environment representation suitable for planning purposes.

In summary, existing approaches either provide accurate state
estimation without reusable uncertainty, or scalable mapping without
accounting for estimation uncertainty.
Methods that partially address these limitations lack representations
that are both probabilistic and efficient for real-time navigation.
This gap motivates representations that preserve \vslam{} uncertainty and
bridge sparse estimation with occupancy-based planning, enabling
uncertainty- and perception-aware navigation in unknown environments.

\subsection{Planning in Unknown Environments}
\label{sec:related_work:planning}

Autonomous planning in unknown environments has gained traction
recently, as it spans different goal functionalities: rapid exploration completion~\cite{bircher2016receding,
  Cieslewski2017, dharmadhikari2020motion,yu2023echo, zhou2021fuel},
increased reconstruction
accuracy~\cite{yoder2016autonomous,schmid2020efficient,
  schmid_fast_2022}, and reliable
localization~\cite{zhang2022exploration, song_learning_2023,
  chen_apace_2024}.

\subsubsection{Rapid exploration}
these methods aim to extend the boundary between known and
unknown space quickly and efficiently.  In \gls{nbv} planning,
candidate viewpoints are sampled and scored to maximize information
gain, while different cost and constraint formulations exist. For example,
\etalcite{Dharmadhikari}{dharmadhikari2020motion} included
\gls{uav} dynamics to achieve feasible and faster exploration.
\etalcite{Bircher}{bircher2016receding} used \gls{nbv} for receding
horizon autonomous exploration.  These approaches often result in
greedy exploration paradigms, leading to dead-ends and failures in
complex scenarios. 

An alternative paradigm is frontier-based exploration, which targets the boundary between known and unknown space. 
\etalcite{Cieslewski}{Cieslewski2017} selected frontiers that minimize change in velocity to maintain high-speed flight, improving exploration efficiency. 
However, the method becomes failure-prone at high speed due to poor estimation performance.
\etalcite{Zhou}{zhou2021fuel} proposed FUEL, an efficient hierarchical framework to perform rapid quadrotor exploration. 
Thanks to a frontier information structure, the system performs three hierarchical steps: global tours, optimal local viewpoints and minimum-time trajectories. Even though it delivers faster trajectories compared to other methods, it assumes perfect state estimation.
\etalcite{Zhang}{zhang_falcon_2024} proposed FALCON, a framework leveraging coverage path guidance for \gls{uav}s with limited \gls{fov} sensors. 
The method stores the environment topology and generates a global coverage path. 
It successfully explores environments using solely on-board state estimation. 
However, the system requires a multimodal sensor stream and does not account for the accuracy of SLAM nor drift, thus degrading the mapping quality over time.

\subsubsection{Accurate reconstruction} 
this type of planning introduces alternative information
gain formulations and is often integrated with sampling-based
methods~\cite{schmid_efficient_2020, schmid_fast_2022,
  moon2025iatigris}.  \etalcite{Palazzolo}{palazzolo2018effective}
proposed a greedy method selecting views to improve 3D uncertainty of
reconstructed landmarks, but did not address localization.
\etalcite{Schmid}{schmid_efficient_2020} first proposed a planning
algorithm to extend a single trajectory tree to perform exploration
and 3D reconstruction.  The method was extended
in~\cite{schmid_fast_2022}, leveraging a learned distribution to
sample new views in partially-observed maps.  Most frameworks are not
favorable for performing tasks in unknown environments.  They require
heuristic-dependent fine tuning to balance exploration and
exploitation. Moreover, they only account for mapping uncertainty,
which often favors novel views which might hinder localization.

\subsubsection{Reliable localization}
\etalcite{Papachristos}{Papachristos2017icra} minimized the
localization uncertainty while exploring the environment.  The method
balances exploration and exploitation with belief space planning.  It
employs multimodal state estimation~\cite{bloesch2017rovio} and
accounts for uncertainty.  The approach was demonstrated in a
visually-degraded cave environment
in~\cite{papachristos_localization_2019}. While the method performed well in
challenging scenarios, it is computationally heavy and requires a
tailored sensor suite.  Moreover, it delivers simple piecewise
linear paths.
\etalcite{Zhang}{zhang2018perception} proposed a kinodynamic RRT* to perform a perception-aware receding-horizon approach that generates a collection of possible trajectories and evaluates them in terms of landmark concentration, collision probability, and distance to the goal.

\etalcite{Bartolomei}{bartolomei_perception-aware_2020} proposed a
kinodynamic A$^*$ to introduce semantic segmentation.  The planned
paths explore reliable features and discard clouds, water and grass.  The
trajectory is smoothed based on a compound cost formulation leveraging
B-splines optimization.  It delivers trajectories balancing smoothness
and co-visibility of as many landmarks as possible.  The method
requires fine-tuning for the cost and assumes perfect knowledge of
both robot pose and map.  Moreover, the approach is tailored for
\glspl{uav} and does not leverage the full 3D state space.  Similarly,
\etalcite{Chen}{chen_apace_2024} introduced a perception-aware
trajectory optimization method for \glspl{uav}.  The system plans in
$\Rthree$ and optimizes a B-spline to select yaws steering the camera
towards higher number of landmarks.  The method leverages ray-casting
and combines a voxel grid and an \gls{esdf}~\cite{han2019fiesta}.
However, it is not suitable for unknown environments, as it assumes a
known sparse map and perfect localization, using estimation merely as
an evaluation tool.

\etalcite{Kondo}{kondo_dynus_2025} devised an uncertainty-aware
planning system to handle dynamic obstacles.  The method implements a
multi-layer planning strategy introducing backup and contingency
trajectories.  It tracks dynamic obstacles through a vision-based
\gls{ekf}, while concurrently estimating an accurate map using
Lidar-Inertial Odometry.  The system adds safety factoring dynamic
obstacle uncertainties leveraging the tracking covariances, achieves
high success rates and delivers high speed trajectories.  However, to
do so, it relies on the use of Lidar, IMU and RGB-D cameras
simultaneously.  The produced trajectories are not perception-aware,
as the planning does not account for the quality of visual features
along the path.

In this work, we focus on a holistic system factoring estimation and planning simultaneously.
We focus on \problemname{} motion, assuming the robot traverses an unknown environment from its current position to a prescribed goal. 
As such, it explores solely necessary sections of the map to successfully reach the goal. 
We propose a novel planning framework to include uncertainty and perception-awareness to deliver safe trajectories even in degraded visual conditions, relying solely on a front-mounted camera.

\subsection{Perception-aware Trajectory Optimization}
\label{sec:related_work:perception_aware_planning}

Trajectory generation delivers optimal trajectories while satisfying constraints, which can be addressed in two ways. 
Hard-constrained methods deliver guarantees on solutions, but hinder computation time and feasibility, while soft-constrained methods introduce cost terms to penalize constraint violations, but cannot guarantee they will never be violated. 

A number of methods use smoothing to satisfy constraints while optimizing for perception quality. 
\etalcite{Murali}{murali2019perception} introduced an optimization method to maximize co-visibility of large numbers of features for \glspl{uav}. 
The same formulation is used by \etalcite{Bartolomei}{bartolomei_perception-aware_2020} and enforce soft-constraints. 
The trajectories are feasible but the process does not scale effectively with map size.
Moreover, these approaches do not account for exploration and assume complete knowledge of the map a priori. 

Similarly, \etalcite{Spasojevic}{spasojevic_perception-aware_2020} proposed a minimum time parametrization preserving features in the camera \gls{fov}.
Most methods perform optimization in $\mathbb{R}^3$ or $\mathbb{R}^4$, not leveraging the full range of motion of the cameras. 
\etalcite{Specht}{spechtLocalizationAware} proposed a data-driven perception-aware cost formulation for space robots in $\SEthree$. 
The method, however, is computationally inefficient and is not suitable for real-time execution.
\etalcite{Chen}{chen_apace_2024} proposed a two step approach optimizing position and yaw, achieving high-speed perception-aware flight thanks to a novel co-visibility cost formulation, but it does not scale with large maps, which are also required to be known a priori.
\etalcite{Tordesillas}{tordesillas_panther_2022} delivered plans that account for the camera \gls{fov} to avoid obstacles, neglecting the actual perception-quality and assuming perfect localization.

In this work, we introduce a novel perception-aware trajectory optimization method on $\SEthree$. 
We leverage the Lie Group and Algebra to efficiently express and handle hard-constraints and uncertainties.
The method is formulated as a \gls{MLE} based on probabilistic visibility of landmarks. 
\methodtrajopt{} handles exploration, occlusions and large maps thanks to ray-casting queries and local submaps. 
Additionally, we steer the camera towards low-uncertainty landmarks rather than just maximizing their number as other methods do.
\begin{table}[t]
\centering
\begin{tabular}{lccccc} 
\toprule
Method & Vision Only SLAM & PP & TO & UA & PA  \\
\midrule
OKVIS2~\cite{okvis2} & \xmark & \xmark & \xmark & \xmark & \xmark \\
OKVISX~\cite{Boche_2025} & \cmark & \xmark & \xmark & \xmark & \xmark \\
ORB-SLAM3~\cite{orb_slam3} & \cmark & \xmark & \xmark & \xmark & \xmark \\
\etalcite{Murali}{murali2019perception} & \xmark & \xmark & \cmark & \xmark & \cmark \\
\etalcite{Bartolomei}{bartolomei_perception-aware_2020} & \xmark & \xmark & \cmark & \xmark & \cmark \\
\etalcite{Specht}{spechtLocalizationAware} & \cmark & \cmark & \cmark & \xmark & \cmark \\
\etalcite{Zhang}{zhang2018perception} & \xmark & \xmark & \cmark & \xmark & \cmark \\
APACE~\cite{chen_apace_2024} & \xmark & \xmark & \cmark & \xmark & \cmark \\
FALCON~\cite{zhang2025falcon} & \xmark & \cmark & \cmark & \xmark & \xmark \\
PANTHER~\cite{tordesillas_panther_2022} & \xmark & \xmark & \cmark & \xmark & \cmark \\
PUMA~\cite{kondo2024puma}& \xmark & \cmark & \cmark  & \cmark & \cmark \\
FUEL~\cite{zhou2021fuel} & \xmark &  \cmark &  \cmark & \xmark & \xmark\\
\etalcite{Papachristos}{papachristos_localization_2019} & \xmark & \cmark & \xmark  & \cmark & \cmark \\
DYNUS~\cite{kondo_dynus_2025} & \xmark & \cmark & \cmark  & \cmark & \xmark \\
\textbf{UNSEEN (Ours)} & \cmark & \cmark & \cmark & \cmark & \cmark\\
\bottomrule
\end{tabular}
\label{tab:algos}
\caption{Comparison between features of various methods used in autonomy stacks. Legend: PP = Path Planning, TO = Trajectory Optimization, UA = Uncertainty Aware, PA = Perception Aware}
\end{table}

\section{Problem Statement and Solution Overview}

\begin{table}[t]
\centering
\caption{Summary of frequently used mathematical notation. }
\label{tab:notation}
\renewcommand{\arraystretch}{1.15}
\begin{tabular}{ll}
\toprule
\textbf{Symbol} & \textbf{Meaning} \\
$\Box^\wedge{}: \Rthree \rightarrow \mathfrak{so}(3)$ & Wedge operator \\
$\Box^\vee: \mathfrak{so}(3) \rightarrow \Rthree$  & Vee Operator\\
$\boxminus$ & Generalized subtraction on Lie Groups \\ 
$B(\cdot, \Box)$ & Geodesic Ball of radius $\Box$ around $\cdot$\\ 

\midrule
$\pose$  & Robot Pose \\
$\hat{\Box{}}$ & Estimated Quantity \\
$\freespace, \obstacle, \unknownspace$ & Free, Occupied and Unknown space \\
$\blockcovariance_{\Box}$ & Marginal Block Covariance of $\Box$ \\
$\residual_{\Box}$ & Residual related to $\Box$ \\
$\blockinformationmat_{\Box}$ & Information Block matrix \wrt $\Box$\\
\hline
$\spacevector \in \Rthree$ &  $3D$ vector \\
$\leafnode$ & Leaf node of the octree \\
$\entityvolume{\Box}$  & Volume of $\Box$  \\
$\localmap_{\Box}$ & Map of $\Box$ \\
\hline
$\tree$ & RRT* Tree \\
$\treenode$ & RRT* Node \\
$\treeedge$ & RRT* Edge \\
$\treepath$ & RRT* Path \\
$\nodecost$ & Cost of RRT* Node \\
\hline
$\angleaxisvector(t) \in \RPthree$ & Angle-axis vector \\
$\translationctrlpoint_i$, $\rotationctrlpoint_i$ & B-spline control points of dimension $\numbercontrolpointstranslation$ and $\numbercontrolpointsrotation$ \\
$\Box \dt$ & Time Derivative \\
$\Bar{\Box}$ & Limit Quantity \\
$\orthobasis{\Box}{i}$ & Orthonormal basis in frame $\Box$ \\
$\constraintmatrix_{\Box}$, $\constraintvector_{\Box}$ & Linear Constraint matrix and known vector \\
\bottomrule
\end{tabular}
\end{table}
\vspace{5mm}
In this work, we address the problem of safe autonomous navigation in unknown, static environments. In particular, we focus on the aspect of \problemname{} motion, which concerns navigating between prescribed start and goal poses through an environment unknown at start.
\subsection{Problem formulation}
Let $\pose_{0}$ and $\posedesfinal$ denote the known start and goal poses of the robot. We assume the robot to be equipped with a forward-looking camera, whose intrinsic and extrinsic parameters are assumed to be approximately known (\eg from factory calibration). In contrast, we do not assume any prior knowledge of the map or any information about the environment before the mission starts.

The navigation problem amounts to finding and executing a collision-free trajectory  $\trajectory$ from start to goal:
\begin{subequations}
\begin{align}
    \trajectory = \{ \poset \vert \forall t \in [t_0, t_F] \}& , \quad \pose (t) \in \SEthree, \\
    \text{subject to:} \nonumber\\
    \label{eq:unseen:traj_def:start}
    \pose (t_0) &= \pose_{0}, \\ 
    \label{eq:unseen:traj_def:end}
    \pose (t_F) &\in B(\posedesfinal, \epsilon_{G}), \\
    \label{eq:unseen:traj_def:obstacle}
    \poset &\in \freespace, \quad \forall t \in [t_0, t_F], \\ 
    \label{eq:unseen:traj_def:dynamics}
    f\bigl(\pose \timeDep, [ \linearvelocity, \angularvelocity]\transpose, u \timeDep\bigr)& \text{ is feasible}, \quad u \timeDep \in \ctrlset\;.
\end{align}
\label{eq:unseen:traj_def}
\end{subequations}
We define $\trajectory$ as the set of continuous poses segments between initial and final times $t_0, t_F$; $B$ represents a geodesic ball centered in $\posedesfinal$ with radius $\epsilon_G$, which is the goal tolerance; $\freespace$ is the free configuration space; $f(\cdot)$ is the system dynamics, which typically depends on $\pose$, linear and angular velocities $\linearvelocity, \angularvelocity$ and control effort $u$, which has to be in the feasible control set $\ctrlset$.
%

The robot relies on onboard \vslam{} for localization and mapping.  Thus, it acts on the estimated trajectory $\estTraj$ and map $\globalmap$.  
We define $\estTraj$ as the set of estimated poses $\estPose$:
\begin{equation}
    \estTraj = \{ \estPose(t) \vert \forall t \in [t_0, t_F] \} , \quad \estPose (t) \in \SEthree.
     \label{eq:unseen:estimated_trajectory}
\end{equation}
The task is successful when the execution of $\estTraj$ leads to
satisfaction of
\eqref{eq:unseen:traj_def:start}-\eqref{eq:unseen:traj_def:dynamics}
for the actual trajectory $\trajectory$.

%
This said, we can formulate the \problemname{} motion planning problem in the following terms:
\begin{subequations}
\begin{align}
    \arg \min_{u(t)} \quad \int_{t_0}^{t_F}& \bigl[\pose\timeDep \boxminus \estPose \timeDep \bigr]\,\mathrm{d}t \\
    \text{subject to:} \nonumber\\
    \estPose_{0} &\notin \unknownspace  \\
    \estposedesfinal &\notin \unknownspace  \\
    \estTraj &\in \estfreespace \\
    \estfreespace &\cap \obstacle = \emptyset \\
    f(\pose \timeDep, [ \linearvelocity, \angularvelocity]\transpose, u \timeDep)& \text{ is feasible}, \quad u \timeDep \in \ctrlset\;,
\end{align}
\label{eq:unseen:planning_problem}
\end{subequations}
\noindent where: $\unknownspace$ is the unknown (\ie unexplored) portion of the environment, $\estfreespace$ is the estimated free space, and $\obstacle$ is the true occupied space.
Based on the problem formulation in~\eqref{eq:unseen:planning_problem}, 
we aim to minimise the deviation between the estimated pose and
the actual pose (computed by the integral of the generalized
subtraction on the Lie groups $\boxminus$)
while respecting the constraints defined in~\eqref{eq:unseen:traj_def}.  In particular,
we enforce that the start and goal states are part of the explored space. 
To ensure safety, we further condition the result so that the
estimated trajectory lies within the estimated free space. To factor
uncertainty, we introduce a condition that prevents occupied space from being
modelled as free.

At runtime,~\eqref{eq:unseen:planning_problem} cannot be solved exactly
due to the unobservability of the required variables.  Instead, we
introduce a perception- and uncertainty-aware end-to-end formulation
that leverages known information to plan $\estTraj$ to find a solution
to~\eqref{eq:unseen:planning_problem}.  These approximations are
detailed in Sec.~\ref{sec:planning}, where we outline our Receding
Horizon Planner.

For clarity, the mathematical notation used throughout this paper is summarized in Table~\ref{tab:notation}. We denote matrices with uppercase letters and block vectors or matrices are in bold.

\subsection{System Overview}
To solve the \problemname{} motion~\eqref{eq:unseen:planning_problem} we introduce a perception- and uncertainty-aware method where the information about uncertainty flows end-to-end. In particular, \methodslam{} receives RGB-D images and computes uncertainties on its estimates, which are concurrently propagated downstream to \methodmap{} and reinjected as regularization terms in the backend optimization. The estimated map $\globalmap$ is fed through \methodmap{}, which updates voxel occupancies accounting for the spatial uncertainty of each estimated landmark. \methodmap{} delivers an efficient memory representation of the map, while allowing to enforce obstacle avoidance and visibility queries. 
\methodplan{} leverages these information to produce perception-aware paths promoting localization accuracy, while delivering free-space representation and local visible submaps, accounting for occlusions and \gls{fov} constraints. Finally, the path is smoothed to avoid venturing in unknown space and promoting visibility of landmarks with low uncertainty. The resulting trajectory is fed in receding-horizon to a low level controller, which will promote state estimation quality by steering the camera towards areas which were confidently estimated before.

\begin{figure*}[t]
\centering
\includegraphics[width=\textwidth]{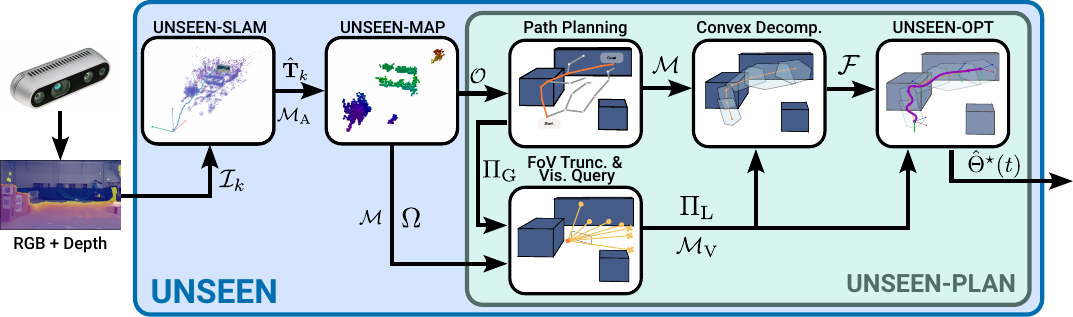}
\caption{Outline of the \method framework. The method relies solely on
  camera images to perform SLAM and Planning in receding horizon. The
  obtained trajectory is then used to produce set-points for the
  low-level controller.}
\label{fig:UNSEEN:Schematics}
\end{figure*}

An overview of the \method{} architecture is shown in
\figref{fig:UNSEEN:Schematics}.
A RealSense D435 captures stereo infrared and monocular RGB images,
performs stereo matching in hardware, and produces a depth image
overlaid on a RGB image $\cameraimage_i$. The UNSEEN-SLAM module
selects and processes keyframes $\keyframes_k$ at \SI{6}{\hertz} to
produce pose estimates and their associated uncertainties, as well as
local feature maps.  The maps are processed in the Octree at \SI{6}{\hertz}.
The planner is triggered every \SI{2}{\second} to produce
global paths into unknown space, which then go through truncation
and smoothing step within our \methodtrajopt{}. 
The trajectory is fed through a low-level controller operating at \SI{100}{\hertz}, delivering thrust commands.

\section{UNSEEN-SLAM \& Occupancy Mapping}
\label{sec:slam&mapping}
Let $\keyframes_k = \{\pose_i\}_{1}^{N}$ be the set of all $N$ keyframes and $\globalmap_k = \{\landmark_j\}_{1}^{M}$ the
global map of $M$ landmarks at time $k$; the state estimates of all keyframes and
landmarks is given by:
\begin{equation}
  \slamstate_k = \left[\keyframes_k,\globalmap_k \right].
\end{equation}
 In common Visual SLAM methods, such as~\cite{orb_slam3}, the \gls{MAP} estimate is
obtained by solving the bundle adjustment over all measurements:
\begin{equation}
    \slamstate_k^* = \arg \max p(\slamstate_k \vert \camerameasurement_k )  = \arg \min \sum_{i \in \keyframes_k} \sum_{l \in \camerameasurement_k} \Vert \cameraresidual{_{i,l}} \Vert_{\camerameasurementcovariance}^2,
\end{equation}
where $\cameraresidual{_{i,l}}$ is the residual of the reprojection error of $\landmark_l$ seen from $\pose_i$, weighted on measurement covariance $\camerameasurementcovariance$.
Then, a visibility graph is built, and old keyframes, i.e., the ones
viewing the active landmarks, are inserted in the optimization as
fixed.

In contrast, in this work, we re-optimize old poses, leveraging covariance information on both poses and landmarks. 
We introduce Tikhonov Regularization~\cite{liu2025tikhonov} residuals on latest estimate of keyframes $\{\hat{\pose}_i\}_{i \in \keyframes_k}$ and landmarks $\{\hat{\landmark}_i\}_{i \in \globalmap_k}$ weighted by their marginal covariances $\posecovariance, \landmarkcovariance$:
\begin{subequations}
\begin{align}
    \residual_{\pose_i} &= \hat{\pose}_i \boxminus \pose_i \\ 
    \residual_{\landmark_i} &= \hat{\landmark}_i - \landmark_i.
\end{align}
\end{subequations}
With slight abuse of notation, we refer to $\keyframes$ and $\globalmap$ as their set of indices, following the same convention as \cite{forster2016manifold}. The resulting \gls{MAP} problem becomes: 
    \begin{align}
    \slamstate_k^* =  \arg \min_{\slamstate} \sum_{i \in \keyframes_k} \sum_{\landmark \in \camerameasurement_k} \Vert \cameraresidual{_{i,l}} \Vert_{\camerameasurementcovariance}^2 \nonumber \\
    + \sum_{i \in \keyframes_k} \Vert \poseregularization{_i} \Vert_{\posecovariance{_i}}^2 
    + \sum_{i \in \camerameasurement_k} \Vert \landmarkregularization{_i} \Vert_{\landmarkcovariance{_i}}^2\;,
    \label{eq:max-a-posteriori}
    \end{align}
which is iteratively solved using the Levenberg-Marquardt method. The solution of~\eqref{eq:max-a-posteriori} is related to the following normal equation: 
\begin{equation}
    \jacobian\transpose \measurementcovariance^{-1} \jacobian \delta\slamstate = \jacobian \measurementcovariance^{-1} \stackedresidual_0\;,
    \label{eq:normal_equation}
\end{equation}
where $\measurementcovariance^{-1}$ is the stacked measurement covariance; $\delta\slamstate$ is the perturbation around the linearization point $\slamstate_0$; $\stackedresidual_0$ is the vector of stacked residuals, also computed in $\slamstate_0$: 
\begin{equation}
    \stackedresidual_0 = \begin{bmatrix}
        \cameraresidual{_{0,0}} (\slamstate_0) \\ \vdots \\ \cameraresidual{_{N,M}} (\slamstate_0)\\ \poseregularization{_0}(\slamstate_0) \\ \vdots \\ \poseregularization{_N} (\slamstate_0)\\ 
        \landmarkregularization{_0}(\slamstate_0) \\ \vdots \\ \landmarkregularization{_N}(\slamstate_0)
    \end{bmatrix}\;,
\end{equation}
and $\jacobian$ is its Jacobian respect to the state $\slamstate$. 

\subsection{Covariance Computation}
\label{sec:covariance-computation}
Since computational efficiency becomes a critical issue
for any SLAM problem on medium-sized areas, we start by identifying on
the left-hand side of~\eqref{eq:normal_equation} the Information
Matrix
\begin{equation}
  \stateinformationmat = \jacobian\transpose \measurementcovariance^{-1}
  \jacobian\; .
\end{equation}
To solve our estimation problem with uncertainty, we need to
efficiently compute the covariance
$\statecovariance = \stateinformationmat^{-1}$. To this end, we
exploit the known structure of the bundle adjustment
problem~\cite{hartley2003multiple}:
\begin{equation}
\stateinformationmat = 
\begin{bmatrix}
\poseposeblockinformationmat & \poselandmarkblockinformationmat \\
\poselandmarkblockinformationmat \transpose & \landmarklandmarkblockinformationmat
\end{bmatrix}.
\end{equation}
During the local bundle adjustment, backend optimization libraries
such as \texttt{g2o}~\cite{grisetti2011g2o} and
GTSAM~\cite{dellaert2017factor} compute the Schur complement $\schurcomplement$ to perform back-substitution and update poses with fewer computation. This is statistically  equivalent to computing the conditional covariance of $\keyframes$ given $\globalmap$ :
\begin{equation}
\schurcomplement = \poseposeblockinformationmat - \poselandmarkblockinformationmat \landmarklandmarkblockinformationmat^{-1} \poselandmarkblockinformationmat \transpose.
\end{equation}

Our method optimizes landmarks and updates the underlying global
map $\globalmap$, which is divided in active and inactive maps
$\activemap$, $\inactivemap$.
In particular, because the $\landmarklandmarkblockinformationmat$ block is block-diagonal and symmetric, its inverse can be computed as follows:
\begin{equation}
\landmarklandmarkblockinformationmat^{-1} = 
\begin{bmatrix}
    \informationmat_{\landmark_0, \landmark_0}^{-1} & \mathbf{0} &\cdots & \mathbf{0} \\
    \mathbf{0} & \informationmat_{\landmark_1, \landmark_1}^{-1} &\cdots & \mathbf{0} \\
    \vdots & \vdots & \ddots & \mathbf{0} \\ 
    \mathbf{0} & \mathbf{0} & \cdots & \informationmat_{\landmark_M, \landmark_M}^{-1}
\end{bmatrix}
.
\end{equation}
Inverting this block of size $3M \times 3M$ goes from complexity
$\mathcal{O}(M^3)$ to a linear complexity $\mathcal{O}(M)$ after
leveraging its structure. This operation is embarassingly parallelisable~\cite{herlihy20120multiprocessor_programming}, with 
a substantial reduction of the computation time.

Once $\schurcomplement$ is available, it is possible to compute the
marginal covariances of the state $\slamstate_k$. Once more, we study
the structure of the resulting covariance matrix, noting we are only
interested in the diagonal blocks:

\begin{equation}
\statecovariance =
\Biggl[
\begin{array}{ccc}
   \cellcolor{green!30} \poseposeblockcovariance &  \poselandmarkblockcovariance \\
  \poselandmarkblockcovariance\transpose & \cellcolor{green!30} \landmarklandmarkblockcovariance
\end{array}
\Biggr]
\end{equation}
%
%
%
Leveraging Schur Decomposition, we reduce unnecessary and expensive
operations and compute the selected blocks~\cite{hartley2003multiple}:
\begin{subequations}
  \begin{align}
    \poseposeblockcovariance &= \schurcomplement^\dagger \\ 
    \landmarklandmarkblockcovariance &= \landmarklandmarkblockinformationmat^{-1} + \landmarklandmarkblockinformationmat^{-1} \poselandmarkblockinformationmat\transpose \poseposeblockcovariance \poselandmarkblockinformationmat \landmarklandmarkblockinformationmat^{-1}
  \end{align}
  \label{eq:LandmarkCovarianceEquation}
\end{subequations}
where the superscript $^\dagger$ represents the Moore-Penrose
Pseudo-Inverse. The computation of the Pseudo-inverse is
$\mathcal{O}(N^3)$, while the Schur decomposition is
$\mathcal{O}(NM^2)$. 

The covariance extracted expresses the accumulated uncertainty,
combining the effects of the measurement uncertainty and of the
feature extraction and matching. In particular, textured areas with
unreliable features will lead to highly uncertain landmark positions,
introducing a covariance increase in the direction of the
surface. This effect will be leveraged to produce occupancy grids for
navigation using solely the optimized landmarks from UNSEEN-SLAM.
\subsection{Covariance Extraction Implementation}
\begin{figure}[t]
    \centering
    \includegraphics[width=\linewidth]{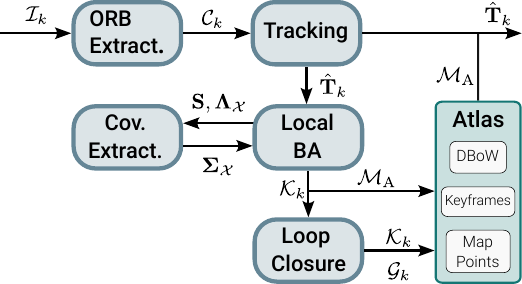}
    \caption{Unseen SLAM multithreaded structure.}
    \label{fig:unseen_slam}
\end{figure}

The covariance extraction described in Sec. \ref{sec:covariance-computation}
is triggered at each local bundle adjustment, but it takes approximately $10$ times the time of a single local bundle
adjustment. For this reason, we perform it asynchronously, as depicted in \figref{fig:unseen_slam}. 
After the feature extraction, the \textit{Tracking} thread estimates poses based on matched features. When new keyframes are selected for insertion, the \textit{Local Bundle Adjustment} thread optimizes the active map $\activemap$ and associated keyframes $\keyframes_k$. 

Once triggered, the \textit{Covariance Extraction} thread updates the covariances of the states, which are then stored in the atlas and reused for further local bundle adjustments. These are updated as soon as they are
available to prevent other threads from being blocked, which would degrade tracking performance. The \textit{Loop Closure} thread reoptimizes the entire active graph without accounting for the uncertainty.

\subsection{UNSEEN Occupancy Mapping}
\label{sec:slam&mapping:occ_mapping}
At runtime, the UNSEEN-SLAM front-end tracks the active map
$ \activemap \subset \globalmap$, which is treated as an
incoming measurement in the form of a point cloud and fed to the \methodmap{}
representation of the environment.  We adopt 
Octomap~\cite{hornung2013Octomap} as the basic structure and introduce
an approach to treat uncertainties and handle local and visible submaps
$\localmap, \visiblemap$.  In particular, we integrate spatial
uncertainty within the probabilistic nature of Octomap, thus
accounting for the spatial extent of measurement uncertainties.

When a new landmark with associated covariance is obtained, its
covariance ellipsoid is computed based on a user-specified threshold
of confidence $\chithreshold^2$.  All cells falling  inside the covariance ellipsoid
are involved in the probabilistic update.  The update is computed by
integrating the PDF over the volume occupied by each voxel. The
resulting integral yields a probability of occupancy which is factored
through the Octomap Bayesian update.

\begin{figure}[t]
    \centering
    \includegraphics{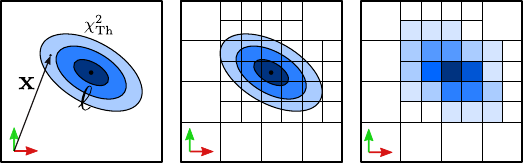}
    \caption{Quadtree toy example of occupancy mapping. \emph{Left:} a new measurement with associated covariance is obtained. \emph{Middle:} a bounding box at the desired resolution of the tree is created. \emph{Right:} each cell is filled with the integral of the Gaussian distribution of the sample and added to the map as log odds.}
    \label{fig:occ-mapping}
\end{figure}

In this work, we propose a new formulation to update the hit
probability for each voxel $n$. A simplified toy example in 2D using
quadtrees is shown in \figref{fig:occ-mapping}.
For each landmark $\landmark_j \in \localmap$, we update all voxels contained in the ellipsoid defined by the following condition: 
\begin{equation}
    (\landmark_j - \spacevector)\transpose \landmarkcovariance{_j} (\landmark_j - \spacevector) \leq \chithreshold^2,
\end{equation}
hence, the likelihood of the voxel containing the estimated landmark
is:
\begin{equation}
    p(\landmark \in \volume_\leafnode) = \int_{\volume_\leafnode} \mathcal{N}(\landmark_j, \landmarkcovariance{_j}) \,\mathrm{d}\spacevector.
\end{equation}
We then define the voxel occupancy probability conditioned on the
landmark as:
\begin{equation}
     p(n \vert \landmark_j) = \, p_{\max} - p_\textup{th}(1 - p(\landmark \in \volume_\leafnode)).
\end{equation}
This value is then fed to the Octomap log-odds formulation as the
\textit{hit probability}. This approach weighs the
occupancy based on the uncertainty of the observation. Therefore, areas with
high confidence are more likely to be considered occupied, while
uncertain sections contribute weakly to the occupied space. This
approach is then effectively coupled with Euclidean clustering, to
remove artifacts and outliers produced by \methodslam{}.
\section{Uncertainty-aware Path Planning}
\label{sec:planning}

Given the current belief of the environment $\globalmap$, i.e., the
global map, we employ RRT* in
$\SEthree$~\cite{sucan2012the-open-motion-planning-library} to sample
the space effectively and consider all possible camera orientations
for path planning, while accounting for discontinuous and non
differentiable costs and constraints at sampling time.  The planner is
both uncertainty-aware and perception-aware, as it considers both the
state of the environment and its uncertainty, as well as provide plans
that facilitates perception and localization tasks down the line. 

\begin{figure}[t]
  \centering
  \subfloat[]{\includegraphics[width=0.48\columnwidth]{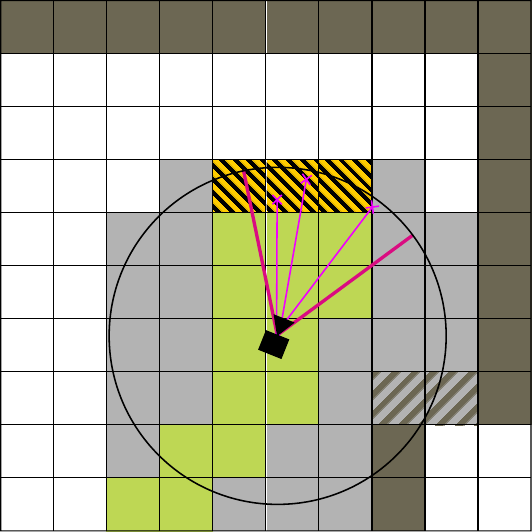}}\quad
  \subfloat[]{\includegraphics[width=0.48\columnwidth]{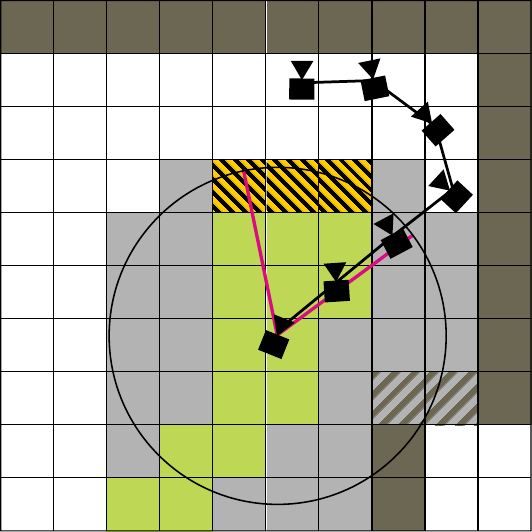}}\\
  \subfloat[]{\includegraphics[width=0.48\columnwidth]{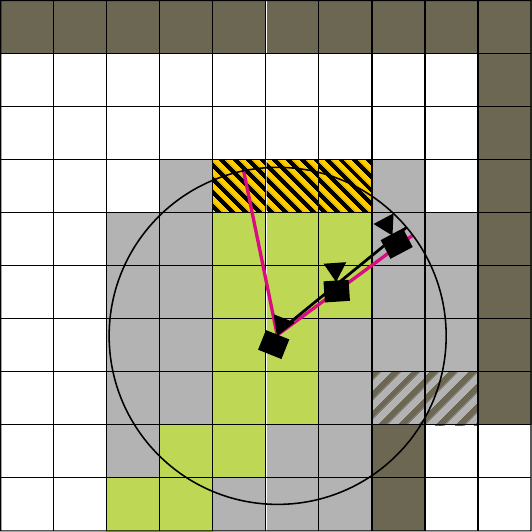}}\quad
  \subfloat[]{\includegraphics[width=0.48\columnwidth]{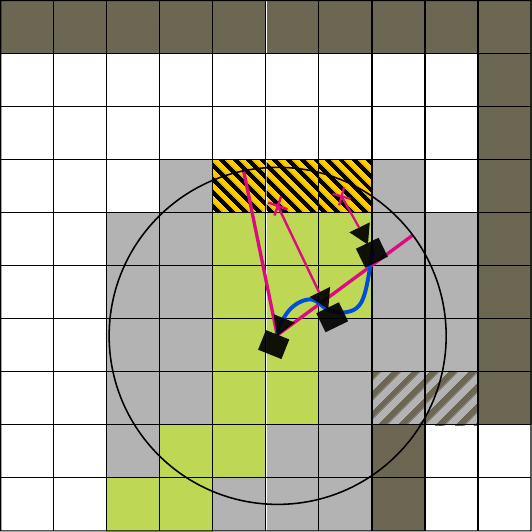}}    
  \caption{UNSEEN Occupancy mapping and planning logic. (a) Textured
    objects (yellow black striped) are seen as landmarks and constitute occupied
    space. Through raycasting, all voxels traversed by the ray
    are set as free (green). All other voxels in the FOV are unknown
    (light gray), untextured obstacles within the FoV (striped gray/olive), 
    untextured obstacles outside FoV (solid olive), 
    and free space (white) are not part of the map and
    are thus unknown space.  (b) A global plan is produced from current
    start to goal, avoiding occupied space. (c) The global plan is
    truncated based on the FOV of the camera. (d) The path is
    constrained so that each camera pose has visible texture and never
    exceeds the \textbf{known and free} (green) space. The path is
    then smoothed to produce a trajectory (blue).}
    \label{fig:planning_logic}
\end{figure}

The planning
step is performed iteratively in a Receding-Horizon fashion, to
account for updates in the map and robot state provided by
\methodslam{}.  The overall planning logic is summarized in
Algorithm~\ref{alg:perception_aware_planning} and
\figref{fig:planning_logic}, and can be split in three steps:
 Global Planning; Field-of-View (\gls{fov}) Truncation and Visibility Checks;
     Convex Decomposition and Smoothing.
\begin{algorithm}[t]
\caption{\methodplan{}}
\label{alg:perception_aware_planning}
\begin{algorithmic}[1]
\REQUIRE Current Map $\globalmap$, current state $\estPose$, goal state $\posedesfinal$, horizon length $H$
\ENSURE Smoothed feasible path segment $\localtreepath$
\STATE \COMMENT{1. Global Planning}
\STATE Initialize RRT* tree $\tree \gets \{\estPose, \posedesfinal\}$

\WHILE{goal not reached}
    \STATE Sample state $\estposerand \in \freespace \cup \unknownspace$ 
    \STATE Find nearest node $\estposenear$ in $\tree$
    \STATE Extend towards $\estposerand$ to get $\estposenew$
    \IF{$\estposenew\in \freespace \cup \unknownspace$ }
        \STATE Evaluate perception-aware metric $\nodecost$
        \STATE Add $\estposenew$ to $\tree$
    \ENDIF
    \STATE Parent Optimization and Rewiring
\ENDWHILE
\STATE Extract global path $\globaltreepath$ from $\tree$ connecting $\estPose$ to $\posedesfinal$
\STATE \COMMENT{2. \gls{fov} truncation}
\STATE \texttt{\gls{fov} Truncation}$(\globaltreepath)\rightarrow \localtreepath$

\FOR{each state $\estPose_i \in \localtreepath$}
    \STATE Free-space decomposition of $\freespace$
    \STATE Visibility check on local map: $\localmap \rightarrow \visiblemap, \visiblevolume$ 
    \IF{$\visiblevolume = 0$}
        \STATE Remove  $\estPose_i$
    \ENDIF
\ENDFOR
\STATE \COMMENT{3. Convex Decomp. and Smoothing}
\STATE \texttt{Convex Decomp.} $(\localtreepath)$
\STATE \texttt{Smooth} $(\localtreepath) \rightarrow \estTraj*$

\RETURN $\estTraj*$
\end{algorithmic}
\end{algorithm}

\subsection{Global Planning}
\label{sec:planning:global_planning}

To provide optimal plans at each iteration, we use the current
knowledge of the entire environment expressed as free and occupied
space within the \method{} Occupancy Map
(Sec.~\ref{sec:slam&mapping:occ_mapping}).  During \textit{Global
  Planning}, we generate a path to the goal ignoring potential
obstacles in the unknown space, \ie assuming
$\unknownspace \subseteq \freespace$.  

From RRT*, we generate a tree
$\tree$ and obtain the optimal path $\globaltreepath$.  During this step, the robot 
is guided toward exploring new areas.  Each sampled state $\posenew$ 
is scored based on an approximated visibility metric:
\begin{equation}
    \nodecost_i  = \frac{\volume_o}{\visiblevolume},
\end{equation}
where $\nodecost_i$ is the cost of node $\treenode_i \in \tree$, $\visiblevolume$ is the volume of visible voxels in the
\gls{fov}, and $\volume_o$ is the occupied volume of voxels in the
local submap $\localmap$.  This metric encourages plans where  the
camera is pointing towards occupied regions, which are more likely to contain
textured areas and reliable landmarks for localization.

\subsection{\gls{fov} Truncation and Visibility Checks}
\label{sec:planning:fov_truncation}

To prevent the robot from entering unknown areas before they are
explored, we verify the global path $\globaltreepath$ from start to
goal. For this purpose, we impose constraints on the tree nodes
$\treenode$ and edges $\treeedge$ to ensure obstacle avoidance, dynamic
feasibility, and compliance with \gls{fov} distance limits. These
constraints remove the overly optimistic assumption that unknown space lies
within free space ($\unknownspace \subseteq \freespace$), which would produce unsafe or unfeasible paths.

Perception and uncertainty are handled through a
\textit{\gls{fov} truncation} step applied after the global plan is
generated. Specifically, the plan is truncated at the boundary between
known and unknown space. Each remaining state is then validated by
checking the visibility of occupied voxels; that is, its \gls{fov}
$\visiblevolume$ must include at least some occupied voxels.

This truncation and constraint validation yields a new local path
$\localtreepath \subseteq \globaltreepath$ that is fully contained in
known free space, never extends beyond the currently visible region, and
always orients the camera toward textured areas. As a result,
$\localtreepath$ is conservative and safe to execute: the robot will not
enter unknown space without exploring it first, and it will not lose
feature tracks due to insufficient texture.

\subsection{Convex Decomposition and Smoothing}
\label{sec:planning:smoothing}

After verifying safety, the discrete path $\localtreepath$ is converted into a smooth, dynamically feasible trajectory on $\SEthree$. This requires extracting local environment information, constructing convex free-space regions, and performing trajectory smoothing under geometric and dynamic constraints.

For each edge $\treeedge$, \methodmap{} provides a local submap $\localmap^\treeedge$ and its visible subset $\visiblemap^\treeedge$. Their unions
\begin{equation}
\localmap = \bigcup_{\treeedge} \localmap^\treeedge, \qquad
\visiblemap = \bigcup_{\treeedge} \visiblemap^\treeedge
\end{equation}
identify the relevant obstacles and occlusions along the path. These become seeds to perform convex decomposition, delivering free-space polytopes $\edgepolytope \subseteq \estfreespace$, and defining the global free-space representation as
\begin{equation}
\polytope = \bigcup_{\treeedge} \edgepolytope.
\end{equation}
These convex sets constrain the trajectory optimization described in Sec.~\ref{sec:trajopt}. Path segments are time-parameterized based on length and dynamic limits, and initialized with a constant nominal velocity $\linearvelocity$.

\subsubsection{Free-Space Extraction and Convex Decomposition}

The map provides access to the estimated free space
\begin{equation}
\estfreespace = \mathbb{R}^3 \setminus \globalmap,
\end{equation}
and to occupied voxels $\estobstacle$. Feasibility of each tree edge is verified by ray-casting: an edge is accepted if the straight segment between its endpoints intersects neither occupied nor unknown voxels.

During smoothing, we again query the occupancy grid to gather obstacle voxels near the path. Exploiting voxel convexity, we extract their vertices and apply the convex decomposition method from~\cite{liu2017decomputil} to generate conservative convex polytopes enclosing free space. These polytopes serve as convex constraints in the optimizer.

\subsubsection{Efficient Map Queries and Visibility}

Each occupied voxel stores the landmarks contributing to its occupancy, enabling efficient obstacle queries. For each node $\treenode \in \localtreepath$, a bounded map query retrieves local obstacles, which are inflated to account for the robot's size.

Visibility is computed using ray-casting, producing the local visible submaps $\visiblemap^\treeedge$. In contrast to prior works~\cite{spasojevic_perception-aware_2020,bartolomei_perception-aware_2020,bartolomei_semantic-aware_2021,chen_apace_2024}, this captures occlusions and sensor \gls{fov} limitations, improving smoothing accuracy while maintaining low computational cost.

\subsubsection{Smoothing}

Finally, the edges of the local path $\treeedge \in \localtreepath$ are interpolated with B-splines for smoothing. Once an initial and final time is appointed for an edge, based on its length and nominal velocity $\linearvelocity$, the segment is optimized to deliver a continuous trajectory in $\SEthree$. Both segments end conditions become the starting boundary conditions for the next. The smoothing is iteratively performed over all edges in the path until a piecewise smooth and continuous trajectory $\estTraj^*$ is obtained, as detailed in Sec.~\ref{sec:trajopt}.

\section{Probabilistic Trajectory Optimization}
\label{sec:trajopt}

We define trajectory optimization as a \gls{MLE}
problem, where we aim to maximize the likelihood of maintaining low-uncertainty landmarks in the field of view. 
Leveraging the visible submap $\visiblemap$, we can query the subset of landmarks that can be seen along an edge $\treeedge$.
We model the visibility of a landmark by defining the camera direction $\cameradirection \in \Spheretwo$ and the landmark's direction as its projected position on the sphere:
\begin{equation}
    \projection(\estPose_t, \landmark_i) = \frac{\estPose_t \landmark_i}{\Vert \landmark_i \Vert }.
\end{equation}
the landmark's covariance is also projected, obtaining: 
\begin{equation}
    \varsigma_{\landmark_i} =\frac{1}{\Vert \landmark_i \Vert} \rotation \landmarkcovariance{_i} \rotation\transpose.
\end{equation}
We can thus formulate the likelihood of observing a landmark $\landmark{_i}$ in the visible map $\visiblemap$ from pose $\estPose_t$ as a Fisher-Bingham distribution, equivalent to a multivariate Gaussian distribution wrapped on the unit sphere $\Spheretwo$ \cite{kume2009fisher}:
\begin{equation}
        p(\cameradirection | \mathbf{T}_t, \landmark_i) \propto (\projection(\estPose_t, \landmark_i) - \cameradirection)\transpose \varsigma_{\landmark_i}^{-1} (\projection(\estPose_t, \landmark_i) - \cameradirection).
\label{eq:fisher-bingham}
\end{equation}

Given \eqref{eq:fisher-bingham} we compute the trajectory by maximizing the likelihood of
observing the landmarks in the visible map $\visiblemap$:
\begin{equation}
  \estTraj^*= \arg \max_{\estTraj} \prod_{t=1}^{N} \prod_{i=1}^{M}
  p(\cameradirection | \estPose_t, \landmark_i) .
\end{equation}
Introducing the log-likelihood of \eqref{eq:fisher-bingham}: 
\begin{align}
    \log \bigl( p(\cameradirection | \mathbf{T}_t, \landmark_i) \bigr) &= \log (e^{-\frac{1}{2}\Vert\projection(\estPose_t, \landmark_i) -
               \cameradirection\Vert_{\varsigma_{\landmark_i}}^2} )  \nonumber\\
               &= - \frac{1}{2}\Vert\projection(\estPose_t, \landmark_i) -
               \cameradirection\Vert_{\varsigma_{\landmark_i}}^2,
\label{eq:log-fisher-bingham}
\end{align}
we finally obtain the optimal trajectory by minimizing \eqref{eq:log-fisher-bingham}:
\begin{align}
  \estTraj^* &= \arg \min_{\estTraj} \sum_{t=1}^{T_f} \sum_{i=1}^{N}
               \frac{1}{2}\Vert(\projection(\estPose_t, \landmark_i) -
               \cameradirection)\Vert_{\varsigma_{\landmark_i}}^2 = \nonumber \\ 
             &= \arg \min_{\estTraj} \optimizationcost(\estTraj, \globalmap) .
               \label{eq:VisibilityCost}
\end{align}

\subsection{B-Spline optimization on Special Euclidean Group}

The infinite-dimensional optimization problem
in~\eqref{eq:VisibilityCost} is solved introducing a smooth
parametrization of the trajectory of the poses:
\begin{equation}
  \estPose \timeDep = 
  \begin{bmatrix}
    \rotation \timeDep & \translation \timeDep \\
    \mathbf{0} & 1
  \end{bmatrix} ,
\end{equation}
using a B-spline representation of the trajectory with the curve
$\optimizationcost : t \in [0, 1] \rightarrow \SEthree$.  To this end,
we first leverage the topological structure of the Lie group
$\SEthree$ as the outer product of $\mathbb{R}^3 \times SO(3)$, thus
opening to the optimization over two separate curves
$(\rotation \timeDep, \translation \timeDep)$. 

As in~\cite{spechtLocalizationAware}, we express the rotational motion
using a composition of an initial orientation $\rotation_0$ and an
angle-axis vector $\angleaxisvector(t) \in \mathbb{RP}^3$, where
$[\angleaxisvector^{\wedge}] \in \mathfrak{so}(3)$ is mapped through
$SO(3)$ exponential map $\exp: \mathfrak{so}(3) \rightarrow SO(3)$.
Consequently, the following a curve:
\begin{equation}
  \rotation \timeDep = \rotation_0 \exp ([\angleaxisvector
  (t)]^{\wedge}) ,
\end{equation}
defines the rotational trajectories of robot's body frame.
Therefore, we can express translation and
rotation separately as:
\begin{align}
  \translation \timeDep &= \sum_{i=1}^{\numbercontrolpointstranslation} \basisfunction_i(t) \translationctrlpoint_i, &
                                                                                                                      \angleaxisvector \timeDep &= \sum_{i=1}^{\numbercontrolpointsrotation} \basisfunction_i(t) \rotationctrlpoint_i,
                                                                                                                                                  \label{eq:splines_w_ctrl_points}
\end{align}
where $\basisfunction_i(t)$ are the B-spline basis functions, while
$\translationctrlpoint_i$ and $\rotationctrlpoint_i$ are the control
points of the translational and rotational B-spline, respectively.

\subsection{Constraints}

To produce feasible trajectories, avoid collisions, and promote
perceptual improvement of the motion, we introduce motion, obstacle and platform
constraints while also keeping the problem solvable in real time.

\subsubsection{Motion Constraints}
To obtain a dynamically feasible trajectory, we enforce conservative bounds on the
robot control effort. Additionally, we actively reduce
motion blur and distortion effects by limiting acceleration and
velocities, for both the linear and angular quantities. Moreover,
in view of the convex hull~\cite{brady1982robot} property of the
B-splines, imposing constraints on the control points
guarantees their satisfaction on the infinite-dimensional continuous
curve.

For the translational trajectory, we define the
velocity and the accelerations as:
\begin{subequations}
    \begin{align}
        \linearvelocity \timeDep   &= \sum_{i=1}^{\numbercontrolpointstranslation -1} \basisfunction_i\timeDep \: \translationctrlpoint_i\dt, &  
        \angleaxisvector\dt \timeDep   &= \sum_{i=1}^{\numbercontrolpointsrotation -1} \basisfunction_i\timeDep \: \rotationctrlpoint_i\dt, \\
        \linearacceleration\timeDep  &= \sum_{i=1}^{\numbercontrolpointstranslation -2} \basisfunction_i\timeDep \: \translationctrlpoint_i\ddt, &
        \angleaxisvector\ddt\timeDep  &= \sum_{i=1}^{\numbercontrolpointsrotation -2} \basisfunction_i\timeDep \: \rotationctrlpoint_i\ddt.
    \end{align}
    \label{eq:constraints:derivatives}
\end{subequations}

For the angular velocity and acceleration, we apply a
small-perturbation approach by mapping the derivatives on the Lie
algebra to dynamic quantities using the left Jacobian
$\leftjacobian(\angleaxisvector) = \leftjacobian$ and its derivative
$\leftjacobian (\angleaxisvector, \angleaxisvector\dt) =
\leftjacobian\dt$, thus yielding:
\begin{subequations}
  \begin{align}
    \angularvelocity &= \leftjacobian  \angleaxisvector\dt, \\
    \angularacceleration &= \leftjacobian\dt   \angleaxisvector\dt + \leftjacobian  \angleaxisvector\ddt.
  \end{align}
\end{subequations}
In order to constrain the orientations, we prescribe limits on the
angular components of $\angleaxisvector$, obtaining a conservative upper
bound vector $\limitangleaxisvector$. This way, we constrain the
trajectory to be enclosed within a Geodesic Ball around $\rotation_0$
as in~\cite{faraci2025reachability}:
\begin{align}
   B( \rotation_0, \Vert \limitangleaxisvector \Vert) &= \Bigl\{ \rotation \in \SOthree \: \vert  \nonumber \\ 
   &\Vert \log(\rotation\transpose \rotation_0)^\vee{} \Vert \leq \Vert \limitangleaxisvector \Vert  \Bigl\}.
\end{align}

Now, we can effectively solve for all higher derivatives limits by
defining the following nonlinear system:
\begin{equation}
    \begin{bmatrix}
        \limitleftjacobian(\limitangleaxisvector) & 0 \\ 
        \limitleftjacobian\dt (\limitangleaxisvector, \limitangleaxisvector\dt) & \limitleftjacobian(\limitangleaxisvector)
    \end{bmatrix} 
    \begin{bmatrix}
        \limitangleaxisvector\dt \\ 
        \limitangleaxisvector\ddt
    \end{bmatrix}
    = 
    \begin{bmatrix}
        \Bar{\angularvelocity} \\ 
        \Bar{\angularacceleration}
    \end{bmatrix}.
    \label{eq:limit_system}
\end{equation}

Despite its nonlinearity, \eqref{eq:limit_system} is lower triangular,
therefore the analytical solution is given by:
\begin{subequations}
\begin{align}
    \limitleftjacobian &= \leftjacobian (\limitangleaxisvector) \\
    \limitangleaxisvector\dt &=  \limitleftjacobian ^{-1} \Bar{\angularvelocity}\\
    \limitangleaxisvector\ddt &= \limitleftjacobian^{-1} (\limitleftjacobian\dt \limitangleaxisvector\dt - \Bar{\angularacceleration}).
\end{align}
\end{subequations}
The main outcome of the proposed approach is that all the needed
quantities can be computed at start time, while the user-defined
constraints can be conservatively applied on the $\SEthree$ trajectory
as follows:
\begin{subequations}
  \label{eq:ConstraintsSE3}
  \begin{align}
    \Vert  \translationctrlpoint_i \Vert_{\infty} \leq \bar{\translation},  &
                                                                              \quad
                                                                              \Vert
                                                                              \rotationctrlpoint_i
                                                                              \Vert_{\infty}
                                                                              \leq
                                                                              \bar{\angleaxisvector}
    ,\\ 
    \Vert \translationctrlpoint\dt_i \Vert_{\infty} \leq \bar{\mathbf{v}} ,  & 
                                                                               \quad
                                                                               \Vert
                                                                               \rotationctrlpoint\dt_i
                                                                               \Vert_{\infty}
                                                                               \leq
                                                                               \bar{\angleaxisvector\dt}
    ,\\
    \Vert \translationctrlpoint\ddt_i \Vert_{\infty} \leq \bar{\mathbf{a}} ,  & 
                                                                                \quad
                                                                                \Vert
                                                                                \rotationctrlpoint\ddt_i
                                                                                \Vert_{\infty}
                                                                                \leq
                                                                                \bar{\angleaxisvector\ddt}, 
  \end{align}
  \label{eq:motion-constraints-long}
\end{subequations}
where \eqref{eq:motion-constraints-long} can be summarized as the following matrix inequality:
\begin{equation}
  \motionconstraintmatrix \controlpointsvector \leq \motionconstraintvector.
\end{equation}

To summarize, this approach introduces some level of conservativeness
and an efficient formulation of constraints as linear
inequalities. The method allows for efficient constraints on both
translational and rotational components of motions and leverages the
properties of B-splines to guarantee constraint satisfaction.  In a
similar way, we can clamp the B-splines to fix boundary conditions on
each edge of the path $\treeedge$.

\subsubsection{Obstacle Avoidance}
Edges $\treeedge$ undergoing smoothing are associated to a visible map
$\visiblemap$ and a free-space convex polytope, resulting in a
selection of non-occluded nearby landmarks and a navigable portion of
free space $\freespace$, as detailed in Sec.~\ref{sec:planning}.  We
constrain each portion of $\estTraj^\star$ to lie in the free-space polytope,
ensuring satisfaction of~\eqref{eq:unseen:traj_def:obstacle}.
Particularly, we leverage the convex hull property of B-Splines to
conservatively guarantee safety against collisions:
\begin{equation}
    \label{eq:trajopt:obstacle_avoidance}
    \freespaceconstraintmatrix \translationctrlpoint_i \leq \freespaceconstraintvector \text{ for } i \in \numbercontrolpointstranslation
\end{equation}
where $\freespaceconstraintmatrix$ and $\freespaceconstraintvector$
are the matrix representing the half-spaces generating the polytope.

\subsubsection{Platform-specific Constraint}
We generate trajectories for a \gls{uav} introducing a differential
flatness constraint.  Our formulation allows for factoring the
specific system dynamics directly in the optimizer formulation, thus
maintaining a level of abstraction and generality.  Since we can define the acceleration along
$\optTraj$ as
in~\eqref{eq:constraints:derivatives}, the specific thrust vector for
our UAV is defined as:
\begin{equation}
  \thrust = \linearacceleration - g \orthobasis{W}{2}
  \label{eq:constraints:specific_thrust}
\end{equation}
where $g$ is the gravitational acceleration.  For the drone to follow
the trajectory $\translation$, it is necessary that:
\begin{equation}
    \rotation  \orthobasis{B}{2} = \frac{\linearacceleration - g
      \orthobasis{W}{2}}{\Vert \linearacceleration - g
      \orthobasis{W}{2} \Vert} = f.
    \label{eq:constraints:z_condition}
\end{equation}
By defining an arbitrary orientation $\rotation_\textup{align}$
satisfying the condition~\eqref{eq:constraints:z_condition} and by
constructing a fictitious frame with $\orthobasis{B}{0}$ in the
forward direction, the constraint satisfaction is ensured iff
$\rotation$ differs from $\rotation_\textup{align}$ by a rotation of
$\psi$ around $\orthobasis{B}{2}$, i.e.
\begin{equation}
  \rotation_{\psi} = \exp \left(\begin{bmatrix}
    0 \\ 0 \\ \psi
  \end{bmatrix}^{\wedge} \right ) ,
\end{equation}
leading to
\begin{equation}
  \rotation_\textup{align}\transpose \rotation =   \rotation_{\psi}.   
\end{equation}

The constraint is then mapped onto the Lie Algebra as:
\begin{equation}
  \angleaxisvector = \log( \big[\rotation_\textup{align}\transpose \rotation \big]  ) = \begin{bmatrix}
    0 \\ 0 \\ \psi
  \end{bmatrix}^{\wedge}.
\end{equation}
Hence, due to the homography between the Lie Algebra and $\Rthree$, we
factor the constraint as a nonlinear equality:
\begin{equation}
  \label{eq:NonLinCons}
    h(\controlpointsvector) = 0,
\end{equation}
which is a function of the control points. Solving the constrained
optimization problem, we can effectively deliver dynamically feasible
trajectories satisfying multiple linear
constraints~\eqref{eq:trajopt:obstacle_avoidance} and one nonlinear
equality constraint~\eqref{eq:NonLinCons}.

\subsection{Gradients}

The proposed B-Spline formulation benefits from efficient gradient
computation with respect to its coefficients as defined
in~\eqref{eq:splines_w_ctrl_points}.  Expressing the optimization
variables as
$\controlpointsvector = \left[ \rotationctrlpoint_0, \hdots,
  \rotationctrlpoint_{\numbercontrolpointsrotation} ,
  \translationctrlpoint_0, \hdots,
  \translationctrlpoint_{\numbercontrolpointstranslation} \right]$, we
can efficiently split the gradient computation over a cost function
$\optimizationcost$ as:
\begin{equation}
    \frac{\partial \optimizationcost}{\partial \controlpointsvector} = \frac{\partial \optimizationcost}{\partial \estTraj} \frac{\partial \estTraj}{\partial \controlpointsvector}.
\end{equation}
We exploit our formulation of the pose: 
\begin{equation}
    \frac{\partial \estTraj}{\partial \controlpointsvector} = \frac{\partial \estTraj}{\partial [\rotation, \translation]} 
    \frac{\partial [\rotation, \translation]}{\partial \controlpointsvector},
\end{equation}
and the B-spline linearity \wrt{} the control points to further
rewrite:
\begin{equation}
    \frac{\partial [\rotation, \translation]}{\partial \controlpointsvector} = 
    \begin{bmatrix}
        \rotation_{0} \frac{\partial \exp([\angleaxisvector]^{\wedge})}{\partial \angleaxisvector} & \mathbf{0} \\ 
        \mathbf{0} & \mathbf{I}
    \end{bmatrix}  
    \frac{ \partial [\angleaxisvector, \translation]}{\partial \controlpointsvector} ,
\end{equation}
and finally introducing the gradient of the B-spline \wrt{} its control points $\controlpointsgradientmatrix$: 
\begin{equation}
  \label{eq:Gradients}
\frac{\partial [\rotation, \translation]}{\partial \controlpointsvector} = 
\begin{bmatrix}
        \rotation_{0} \leftjacobian & \mathbf{0} \\ 
        \mathbf{0} & \mathbf{I}
    \end{bmatrix}  
\begin{bmatrix}
        \controlpointsgradientmatrixrotation & \mathbf{0} \\ 
        \mathbf{0} & \controlpointsgradientmatrixtranslation
    \end{bmatrix}  .
\end{equation}

\subsection{Optimal Control Problem Formulation}

With the analytic reformulation introduced within this section, we
propose a finite-dimensional relaxation of \problemname{} motion
planning~\eqref{eq:unseen:planning_problem} as:
\begin{subequations}
  \begin{align}
    \controlpointsvector^* = &\arg\min_{\controlpointsvector} \optimizationcost ( \estTraj(\controlpointsvector), \globalmap) \\ 
    \text{subject }& \text{to} \nonumber \\ 
                             &\motionconstraintmatrix \controlpointsvector \leq \motionconstraintvector, \\ 
    \bigwedge_{i=1}^{\numbercontrolpointstranslation} &\freespaceconstraintmatrix \translationctrlpoint_i \leq \freespaceconstraintvector, \\
                             &h(\controlpointsvector) = 0 ,     
  \end{align}
\end{subequations}
that, using the expressions of the gradients in~\eqref{eq:Gradients}
can be solved numerically via SLSQP using \texttt{NLopt}.

\subsection{Validity of B-Spline Representation on \texorpdfstring{$SO(3)$}{SO(3)}}
\label{sec:trajopt:so3_spline}

In this work, we parameterize the rotational trajectory using a
B-spline in the Lie algebra $\SOthreealgebra$, mapped to $\SOthree$
through the exponential map. This representation provides smooth
interpolation and enables direct enforcement of motion constraints in
a vector space. While cumulative B-splines on Lie
groups~\cite{sommer2020efficient} offer a fully group-consistent
formulation based on recursive compositions and \gls{bch} \cite{rossmann2006lie} corrections, our setting does not require such
machinery. The trajectory optimization explicitly constrains angular
velocity and acceleration, ensuring that consecutive control points
increments remain well within the injectivity radius of the
exponential map. As a result, rotations stay in a locally Euclidean
neighborhood where the logarithm is unique and linear interpolation in
$\mathfrak{so}(3)$ remains accurate.

These bounded increments eliminate the need to involve an incremental structure, allowing us to rely on properties of convex neighborhoods and local diffeomorphisms without additional complexity. 
Moreover, this approach significantly reduces computational overhead, making it suitable for real-time applications while maintaining sufficient accuracy for the constrained maneuvers considered in this work.

\section{Experimental Results}
\label{sec:experimental_results}

To evaluate the contributions of \method{}, we test its different
components on a robotic platform both in isolation to benchmark their performance, as well as in full pipeline.
For a repeatable and fair
comparison with the state-of-the-art, some of the tests were also carried out in simulation.

\subsection{Experimental Setup}
\label{sec:experimental_results:setup}
The experimental setup consisted of different scenarios specifically
designed to test \method{} by presenting little reliable texture in
the environment. A $11\times 20 \times 10 \text{m}^3$ flight arena was
filled with textureless cardboard boxes, some with AprilTag markers
applied to them only on certain areas.  The AprilTags were used solely
to artificially introduce well-established and recognizable textured
areas in the map, no specific AprilTag detector was used for any
purpose. 

For safety reason, the drone had a software-level constraint to move within a defined safe volume of $6 \times 12 \times 2 \text{m}^3$
inside the arena, with a maximum speed of $1.5 \text{m/s}$, as well as an RC-controller setup to manually operate it in case of emergency.  The software constraints
were iteratively verified and violation would lead to termination of
the flight.

\begin{figure*}[t]
\centering
\subfloat[][Easy]{
\includegraphics[width=0.25\textwidth]{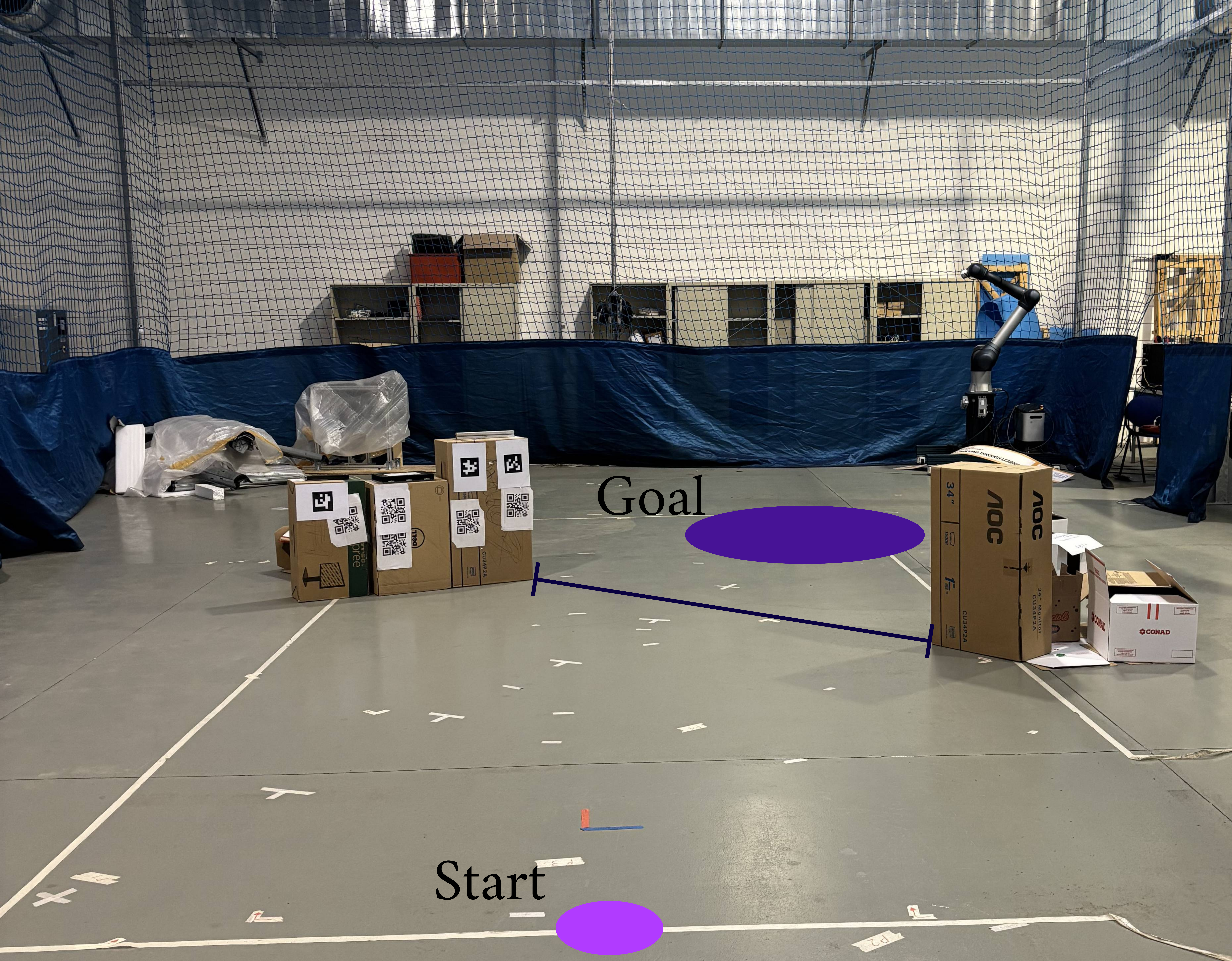}}
\subfloat[][Medium]{
\includegraphics[width=0.25\textwidth]{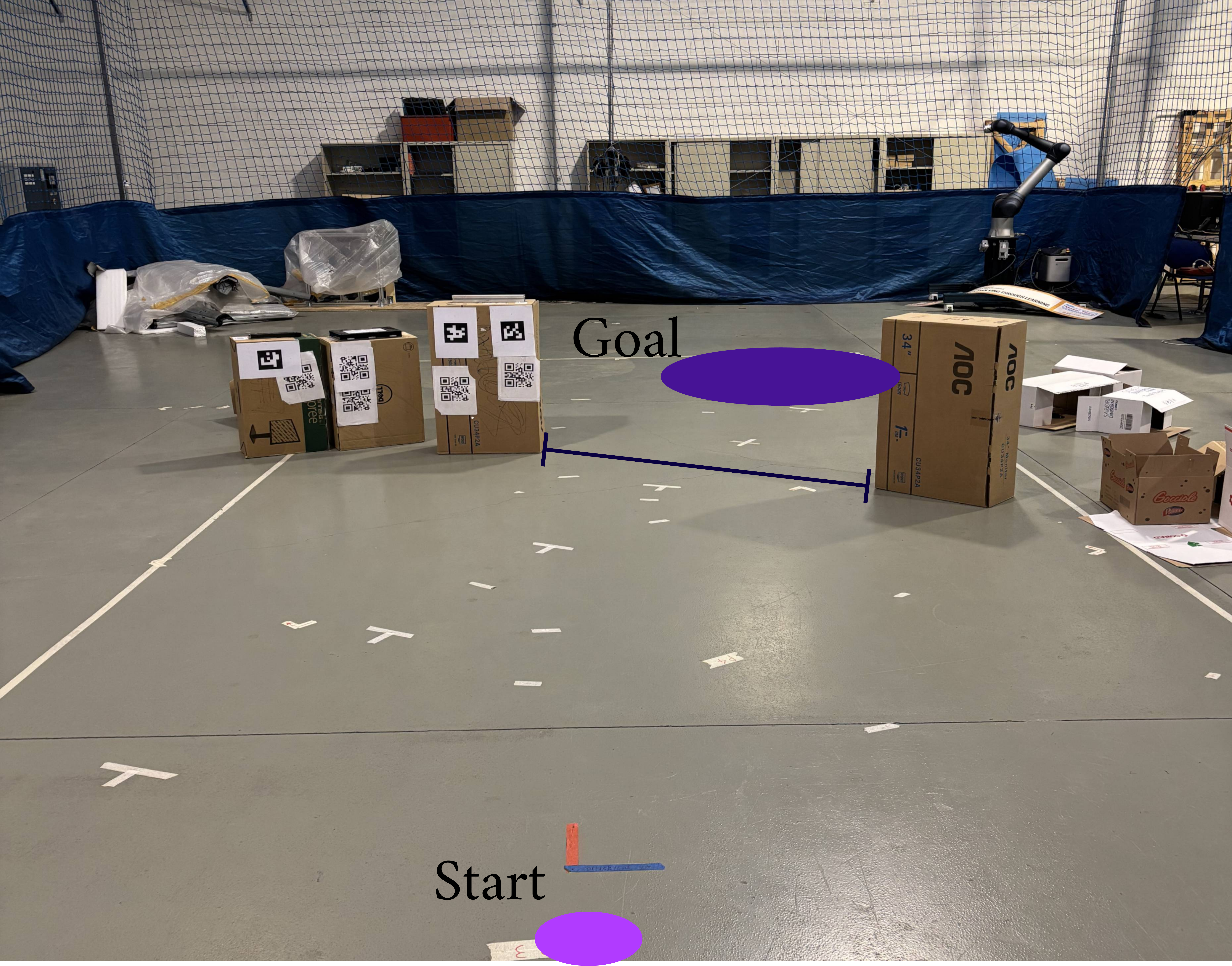}}
\subfloat[][Hard]{
\includegraphics[width=0.25\textwidth]{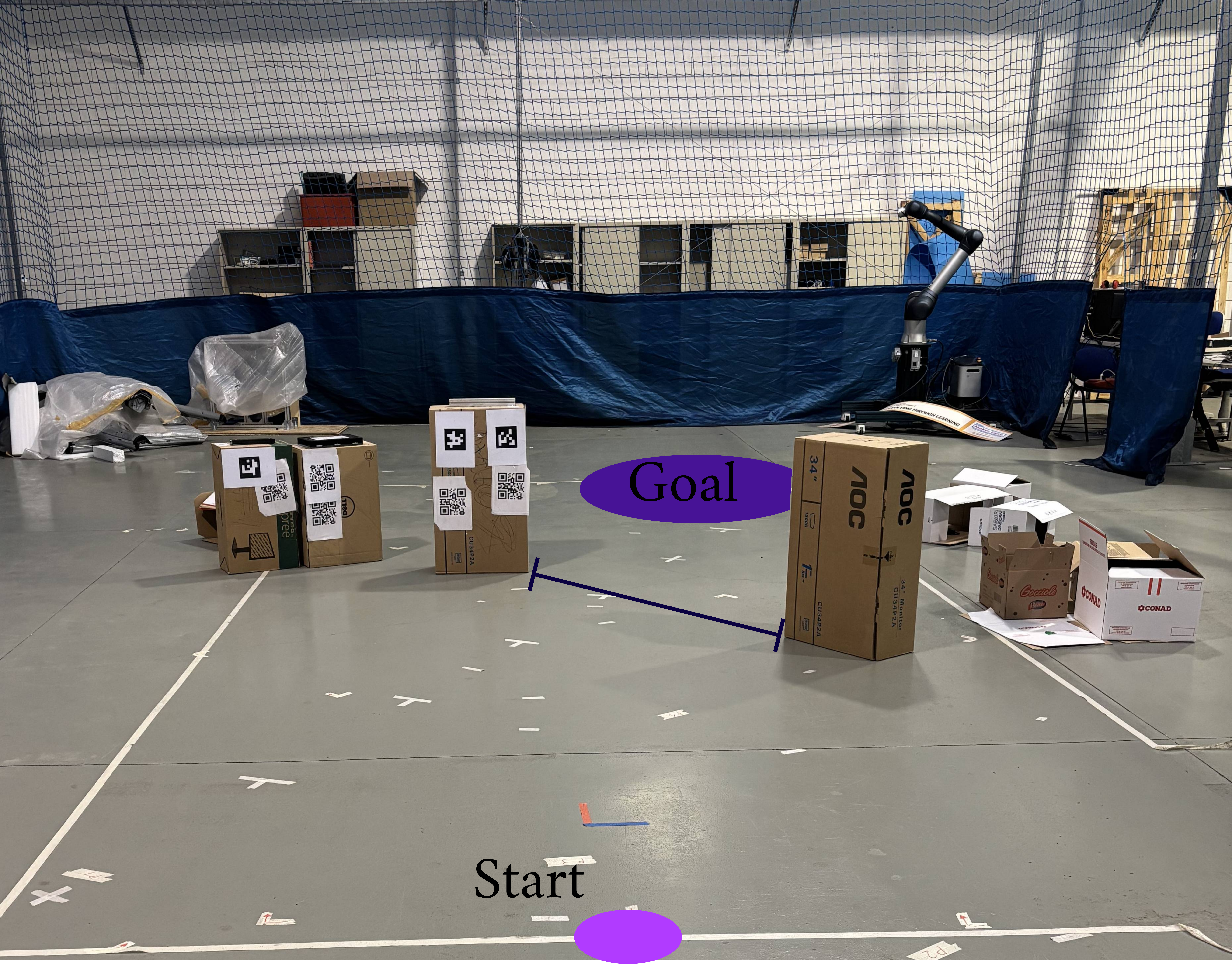}}
\caption{Experimental scenarios.}
\label{fig:experimental_setup}
\end{figure*}

The boxes were arranged in three different configurations to establish
difficulty levels for \problemname{} task: Easy, Medium, and Hard.
The level of difficulty is determined by the size of the passage the
drone had to traverse between boxes, as well as the amount of the
field of view occupied by them.  \figref{fig:experimental_setup} shows
the three scenarios used in the experiments.

\begin{figure}[t]
  \centering
  \includegraphics[width=0.75\columnwidth]{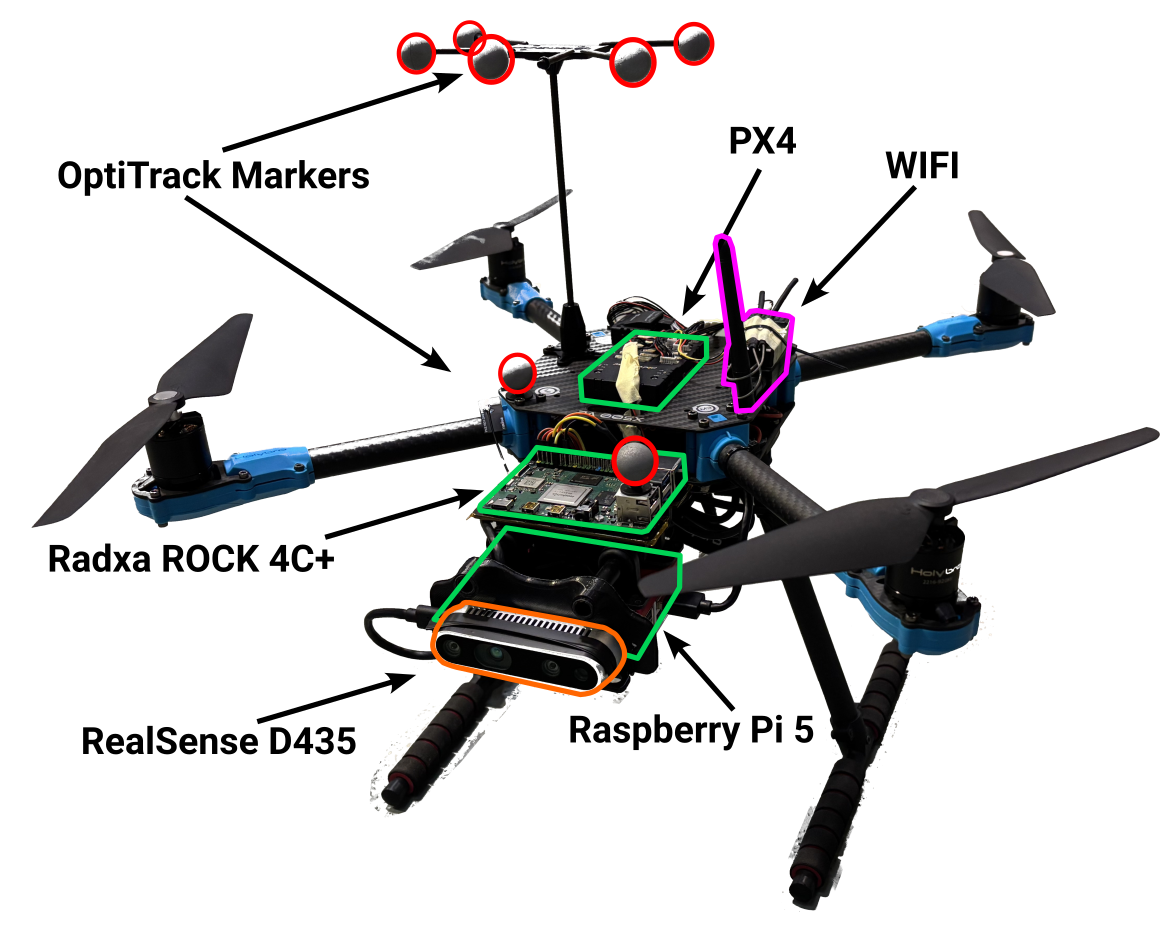}
\caption{Experimental platform: a Holybro X500 v2 equipped with an
  Intel Realsense D435 RGB-D camera and two low-power companion
  computers.}
    \label{fig:experimental_platform}
\end{figure}

The experimental platform used is a Holybro X500 v2, shown in
\figref{fig:experimental_platform}.
It is equipped with two low power single board companion computers: a
\texttt{Raspberry Pi 5} to run the \texttt{Intel RealSense D435}
camera driver and a \texttt{Radxa ROCK 4C+} to interface with the PX4
flight controller.  The \method{} pipeline was executed on a
\texttt{Intel i7} laptop communicating via Wi-Fi with the companion
computers to receive camera streams at 6 Hz and send back waypoints
via \texttt{ROS2}.  A manual controller and the
\texttt{QGroundControl} application were used as backups to ensure
safety, and provide telemetry. 

The ground truth trajectory was provided by an \texttt{OptiTrack}
motion capture system, which provided the full pose of the robot with
millimeter precision.  However, the motion capture system was only
used for evaluation purposes and no information was fed to the onboard
estimation or planning systems, except for the software-level fencing
previously described. A total of $10$ flights were performed. The
pipeline relied solely on the perceptual cues from the RGB-D images
from the \texttt{Realsense D435}.

\subsection{\methodslam{} Testing}
We demonstrate the effectiveness of \methodslam{} in navigation tasks
by comparing its performance to the baseline ORB-SLAM3.  The two
methods share the same configuration layout, thus they ran on the
exact same set of parameters, reported in
Tab.~\ref{tab:slam_parameters}.
\begin{table}[t]
\centering
\caption{SLAM parameters used for both \methodslam{} and ORB-SLAM3.
Width, Height, Rate, Number of features per frame, Minimum and Maximum depth, Scale factor and Number of pyramid levels.}
\label{tab:slam_parameters}
\begin{tabular}{c|c|c|c|c|c|c|c}
\toprule
$\cameraimage$ w. &
$\cameraimage$ h. &
$\cameraimage$ r. &
$\camerameasurement$ &
$d_{\min}$ &
$d_{\max}$ &
$s$ &
$\# lev.$ \\
\midrule
$424$ 
& $240$ 
& $6$ Hz 
& $600$ 
& $0.5$ m 
& $5.0$ m 
& $1.2$ 
& $8$ 
\end{tabular}
\end{table}
The requirements of \methodmap{} and \methodplan{} do not allow other
\vslam{} method to be employed without losing significant
capabilities. Therefore, we gathered a small dataset of recorded
trajectories to test different \vslam{} methods offline.

Table~\ref{tab:slam_ape_rpe} summarizes the results of the comparison
between \methodslam{} and ORB-SLAM3 in terms of \gls{APE} and
\gls{RPE}.
\begin{table}[t]
\centering
\caption{Absolute and Relative Pose Error comparisons between ORB-SLAM3 (OS3)~\cite{orb_slam3} and \methodslam{}(Ours). 
E = Easy, M = Medium, H = Hard}
\label{tab:slam_ape_rpe}
\setlength{\tabcolsep}{4pt}    
\renewcommand{\arraystretch}{0.98}
\begin{tabular}{l c| cc| cc| cc| cc}
\toprule
 & & \multicolumn{2}{c}{\textbf{ATE [m]}} & \multicolumn{2}{c}{\textbf{ARE [deg]}} &
 \multicolumn{2}{c}{\textbf{RTE [m]}} & \multicolumn{2}{c}{\textbf{RRE [deg]}}  \\
\cmidrule(lr){3-4} \cmidrule(lr){5-6} \cmidrule(lr){7-8} \cmidrule(lr){9-10}
\textbf{D} & \textbf{R} & \textbf{OS3} & \textbf{UN} &  \textbf{OS3} & \textbf{UN} & \textbf{OS3} & \textbf{UN} & \textbf{OS3} & \textbf{UN} \\
\midrule
\multirow{3}{*}{E} & 1            & 0.358   & \cellwin{0.332}   & \cellwin{0.17}    & 0.30            & 0.301             & \cellwin{0.289}       & \cellwin{0.10}   & 0.11\\
                   & 2            & 0.542   & \cellwin{0.266}   & 0.28              & \cellwin{0.26}  & 0.234             & \cellwin{0.218}       & \cellwin{0.08}   & 0.09\\
                   & 3            & 0.397   & \cellwin{0.382}   & \cellwin{0.16}    & 0.18            & \celldraw{0.212}  & \celldraw{0.212}      & 0.06             & \cellwin{0.05}\\
\midrule            
\multirow{4}{*}{M} & 4            &  0.357  & \cellwin{0.205}   & \cellwin{0.14}    & 0.16            &  0.166            & \cellwin{0.165}       & \celldraw{0.05}   & \celldraw{0.05} \\
                   & 5            & 0.515   & \cellwin{0.424}   & \cellwin{0.11}    & 0.38            &  \cellwin{0.304}  & 0.312                 & \cellwin{0.07}   & 0.08 \\
                   & 6            & 0.537   & \cellwin{0.450}   & \cellwin{0.16}    & 0.18            &  \cellwin{0.332}  & 0.358                 & \cellwin{0.09}   & 0.10  \\
                   & 7            &  0.402  & \cellwin{0.368}   & \cellwin{0.28}    & 0.34            &  0.313            & \cellwin{0.303}       & 0.16             & \cellwin{0.13} \\
\midrule            
\multirow{3}{*}{H} & 8            & 0.434   & \cellwin{0.360}   & \cellwin{0.05}    & 0.06            &  0.159            & \cellwin{0.143}       & \celldraw{0.03}   & \celldraw{0.03} \\
                   & 9            &0.397    & \cellwin{0.261}   & \cellwin{0.05}    & 0.07            & 0.159             & \cellwin{0.145}       & \cellwin{0.03}   & 0.04            \\
                   & 10           & 0.433   & \cellwin{0.328}   & 0.38              & \cellwin{0.37}  & 0.198             & \cellwin{0.187}       & \celldraw{0.04}   &  \celldraw{0.04}      \\
 \midrule
\multicolumn{2}{l}{\textbf{Mean}} &  0.408  & \cellwin{0.368}   &   \cellwin{0.20} &  0.23          & 0.241             & \cellwin{0.235}       & \celldraw{0.08}  & \celldraw{0.08}  \\
\multicolumn{2}{l}{\textbf{Cov.}} &  0.021  & \cellwin{0.014}   &  \celldraw{0.01} & \celldraw{0.01}          & \cellwin{0.004}   & 0.005                 & 0.01             & \cellwin{0.00}   \\
\bottomrule
\end{tabular}
\end{table}
The results show the \methodslam{} ability to operate online,
demonstrating how the asynchronous covariance extraction
(Sec.~\ref{sec:slam&mapping}) does not lock nor hinder the tracking
performance.

Moreover, \methodslam{} uncertainty-aware capabilities improve the
absolute and relative translation error, introducing a $9\%$ and $2\%$
improvement within navigation tasks, respectively.  The table shows
that the introduction of the regularization does degrade performance
in terms of absolute rotation error, although it remains comparable to
ORB-SLAM3 with a $0.03^\circ$ increase on average.  Within the scope
of navigation, we argue that translation accuracy is of higher
importance and impacts the success of the task more significantly than
orientation accuracy.

\subsection{\methodplan{} Testing}

\begin{figure*}[t]
    \centering
    \begin{minipage}{\columnwidth}
    \centering
      \includegraphics[width=0.8\linewidth]{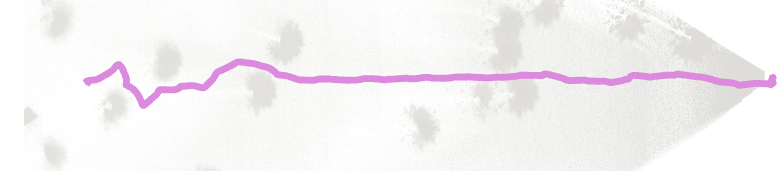}
      {\small(a)}
    \end{minipage}%
    \begin{minipage}{\columnwidth}
    \centering
        \includegraphics[width=0.8\linewidth]{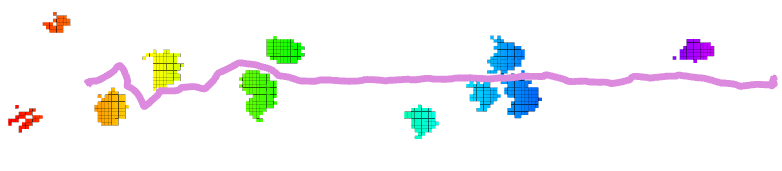}
        {\small(b)}
    \end{minipage}
    \begin{minipage}{\columnwidth}
    \centering
      \includegraphics[width=0.8\linewidth]{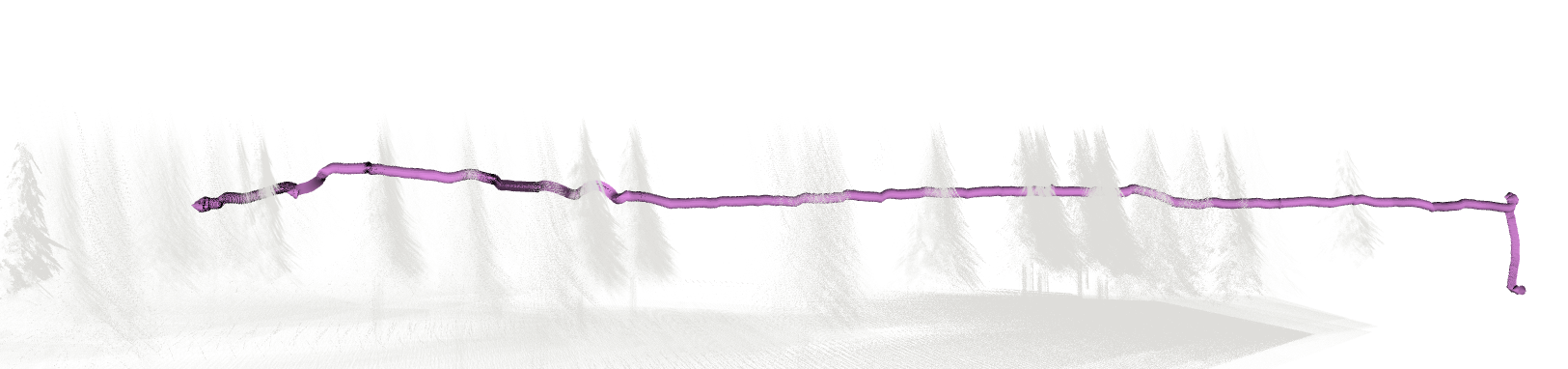}
      {\small(c)}
    \end{minipage}%
    \begin{minipage}{\columnwidth}
    \centering
      \includegraphics[width=0.8\linewidth]{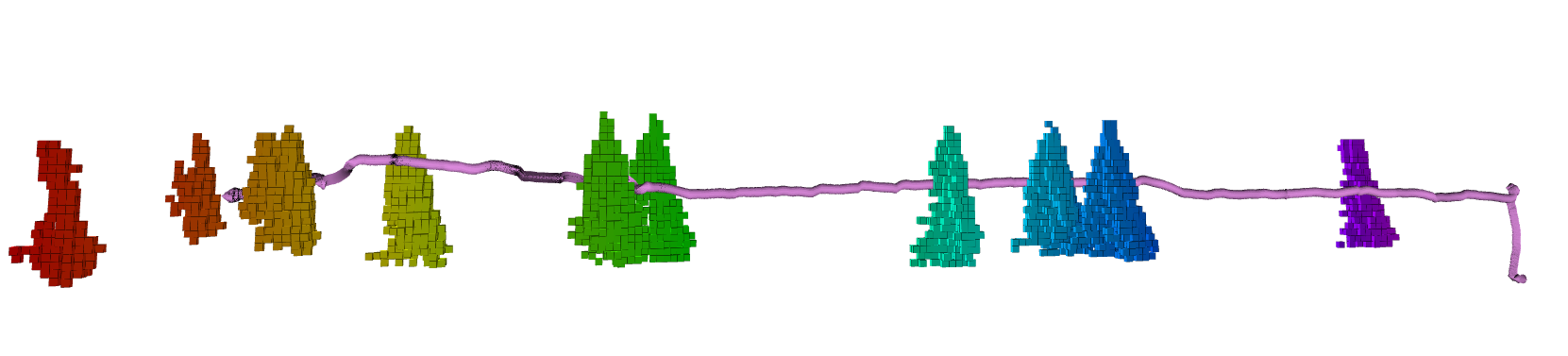}
      {\small(d)}
    \end{minipage}
    \caption{Resulting path delivered by \method{} in the
      \textit{Forest} environment. (a) and (c) show the ground truth
      as the accumulated point clouds. (b) and (d) show the associated
      UNSEEN Occupancy Map generated during the flight.}
    \label{fig:res:forest_planning_sim}
\end{figure*}

For the sake of reproducibility and benchmarking, the planning
component of \method{} has been evaluated qualitatively in simulation.
In particular, the method was tested in a \textit{Forest Environment},
consisting of a $20 \mbox{ m} \times 20 \mbox{ m} \times 5 \mbox{ m}$
area populated with $50$ randomly placed trees of $0.3$~m radius and
heights between $3$~m and $5$~m.  The environment is designed to mimic
a natural outdoor scenario, with a high density of textured obstacles,
which however present a challenge in terms of feature matching and
place recognition, due to the repetitive patterns of tree bark and
foliage.  The starting and goal positions were placed at opposite
sides of the environment, requiring the planner to navigate through
the dense forest while avoiding collisions. To overcome the issues of
photorealism in simulation, we simulate the output of \methodslam{} by
generating a noisy pose estimate and an uncertain map, based on the
output of a simulated RGB-D camera. The results of the simulated path
and \methodmap{} reconstruction in the \textit{Forest Environment}
are shown in \figref{fig:res:forest_planning_sim}. For these experiments, the confidence threshold $\chithreshold$ is set at 95\% confidence, thus: $\chithreshold^2 = \chi_{0.95} = 0.352$.

As noted in Sec.~\ref{sec:related_work}, no other method performs
\problemname{} planning relying solely on visual information with
estimation in the loop.  Therefore, it is not possible to perform a
direct comparison on the robotic platform.  Hence, we test \methodtrajopt{} in simulation 
on a simpler scenario which the same assumptions found in literature.
In particular, the environment is assumed to be
already known at planning time and no exploration is necessary.
Furthermore, an occupancy map is available and no replanning is
required.  \methodtrajopt{} operates with ground truth poses and
performs simple point-to-point motion.

\begin{figure}[t]
  \subfloat[Gazebo]{\includegraphics[height=3.2cm]{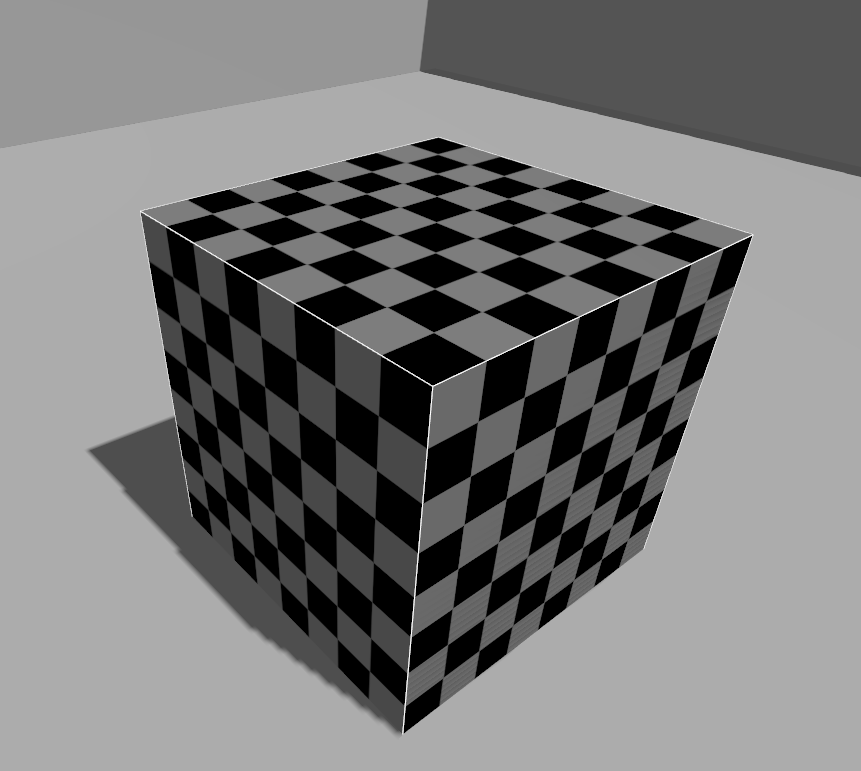}}
  \subfloat[Map]{\includegraphics[height=3.2cm]{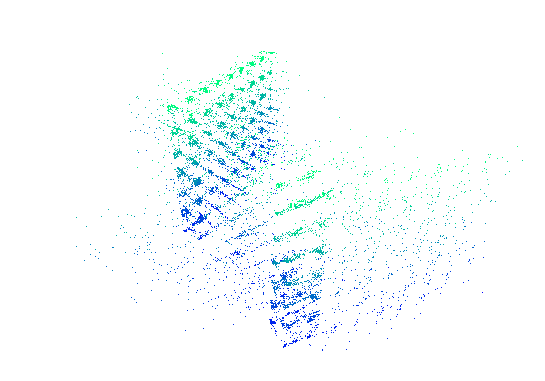}}
    \caption{\textit{Cube Benchmark} scenario. Only the cube has been
      resolved, the reconstruction of the walls is not possible due to
      lack of texture. (a) Point cloud map from RGB images. (b)  Depth
      point cloud.}
    \label{fig:res:cube_benchmark}
\end{figure}

We evaluate the planning pipeline and measure its perception-aware
capabilities by designing the \textit{Cube Benchmark} scenario,
depicted in \figref{fig:res:cube_benchmark}.
The robot is tasked with navigating around a
$5 \mbox{ m} \times 5 \mbox{ m} \times 5 \mbox{ m}$ cube placed in the
center of a $15 \mbox{ m} \times 15 \mbox{ m} \times 5 \mbox{ m}$ room
with textureless walls. As the only visible texture is provided by the
cube, trajectories must maintain the cube in the \gls{fov} view at all
times to avoid loss of tracking and failure of the navigation task.
As previously stated, no estimation is part of the loop for this test. However,
\methodslam{} is employed to provide an evaluation of the final
trajectory with respect to the ground truth.

\begin{figure}
    \centering
    \includegraphics[width=\columnwidth]{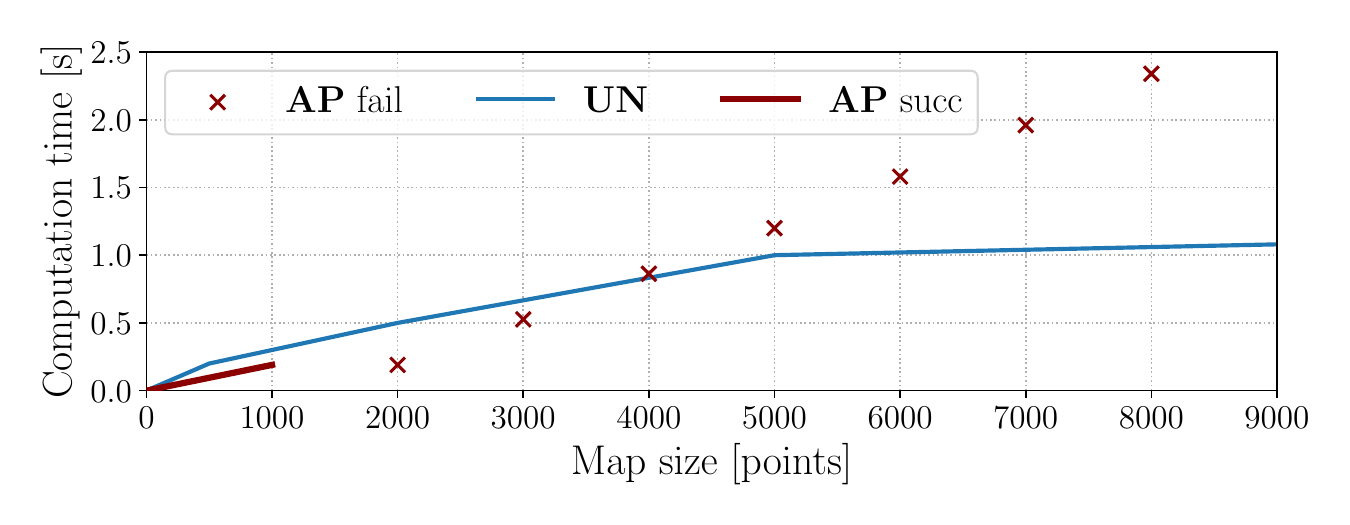}
    \caption{Computation times against map size for the Cube benchmark.}
    \label{fig:computation-time}
\end{figure}

With these matching assumptions and scenario, we compare UNSEEN-Plan
with the most similar and current state-of-the-art method, APACE~\cite{chen_apace_2024}.  First, we obtain a map by
simulating a camera trajectory around the cube and feeding the RGB-D
images to \methodslam{} (see \figref{fig:res:cube_benchmark}).  The
resulting map $\globalmap$ is fed to both planners to compute a path
around the cube. 
Table\ref{tab:apace_vs_unseen} reports success/failure against map size, up to 40000 elements, while \figref{fig:computation-time} shows the computation times. \methodplan{} achieves a bounded computational time due to the submap extraction and visibility query strategy described previously, while APACE's reaches increases exponentially. 
On 40000 map points, APACE fails after 22 seconds, while \methodplan{} succeeds in less than 2.0 seconds.
\begin{table}[t]
\centering
\caption{Success of PAW methods with different map sizes with time
  elapsed before failure or success}
\label{tab:apace_vs_unseen}
\begin{tabular}{l|ccccc}
\hline
\multirow{2}{*}{\textbf{Method}}
 & \multicolumn{5}{c}{\textbf{Map Size (points)}} \\
& 500 & 2000 & 5000  & 15000 & 40000 \\
\hline
APACE~\cite{chen_apace_2024} &  \celldraw{}\cmark  &  \xmark   & \xmark  &  \xmark & \xmark   \\
UNSEEN  &  \cellwin{}\cmark  &  \cellwin{}\cmark & \cellwin{}\cmark &  \cellwin{}\cmark  & \cellwin{}\cmark    \\
\hline
\end{tabular}
\end{table}

APACE struggles with handling large maps, as the system is not
designed to efficiently deal with a large number of points.  In
contrast, \method{} embedded submap and visibility query allows for
easy handling of large number of points. Finally, we report the
performance of both methods. While APACE delivers consistently the
same trajectory, \method{} showcases more variable outcomes in the
vicinity of a mean trajectory.  However, \method{} trajectory
consistently outperforms APACE in terms of \gls{APE} even when
deployed on the same map and evaluated with the same algorithm.

\begin{figure}[t]
    \centering
    \begin{minipage}{0.5\columnwidth}
      \includegraphics[width=0.8\linewidth]{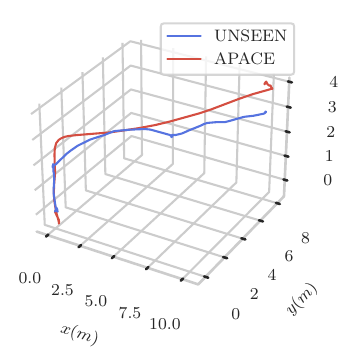}
      {\small(a)}
    \end{minipage}
    \begin{minipage}{0.5\columnwidth}
      \includegraphics[width=0.8\linewidth]{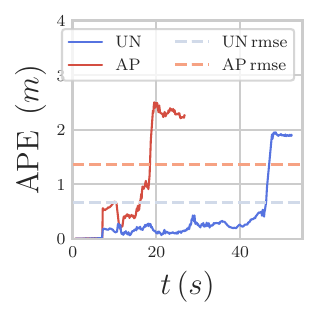}
      {\small(b)}
    \end{minipage}
    \caption{Comparison of APACE and UNSEEN-Plan trajectories in the
      \textit{Cube Benchmark} scenario. (a) 3D Trajectories. (b)
      \gls{APE} over time.}
    \label{fig:apace_unseen_traj}
\end{figure}

As a further analysis, we may notice that while APACE plans one single
smoothed B-spline over the path, \method{} delivers a $\mathcal{C}^2$
piecewise polynomial trajectory, which constrains the robot to
traverse waypoints specifically optimized to improve perceptual
capabilities by the global planner.  Moreover, APACE promotes aggressive
maneuvers while \method{} introduces velocity and acceleration
constraints, which ultimately aid perception and estimation
guaranteeing low motion blur and consistent and reliable tracking
across frames.  The results are shown in
\figref{fig:apace_unseen_traj}.
%

\subsection{Perception-Aware Planning Testing}

\begin{figure}[t]
  \centering
  \includegraphics[width=0.8\linewidth]{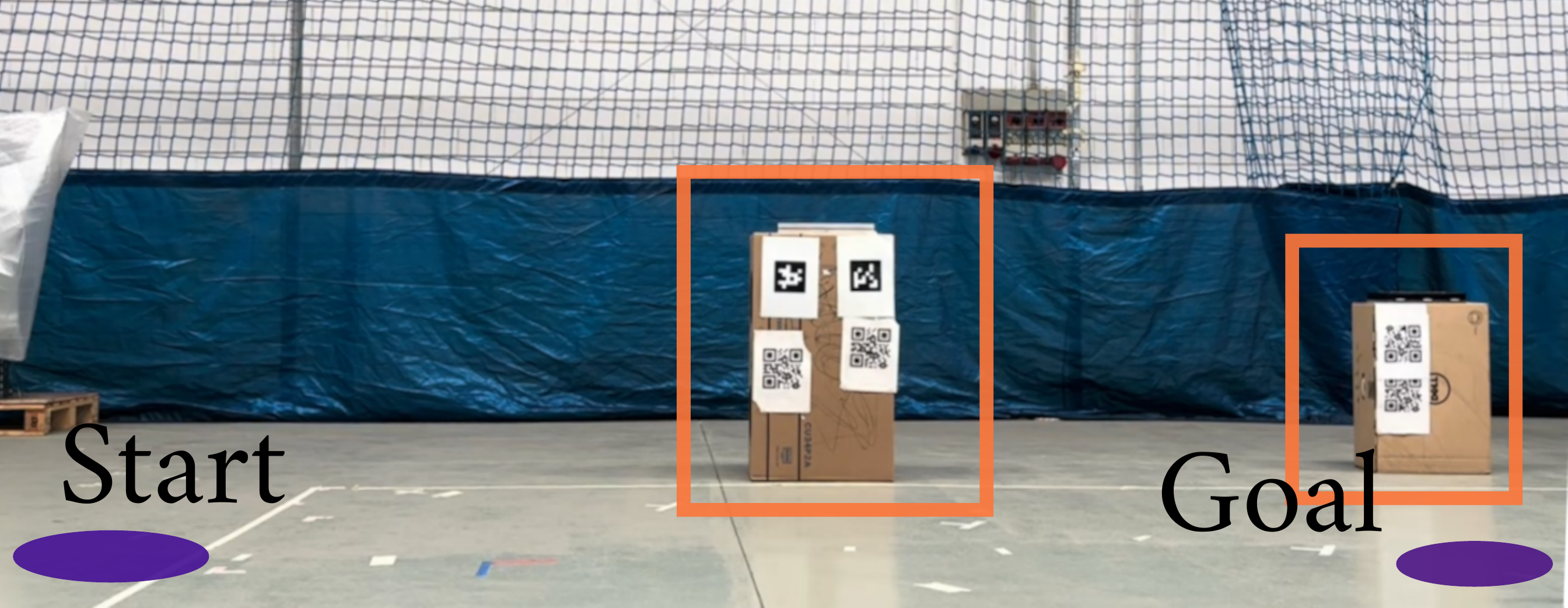}
  \caption{Setup for the PAW vs PAG experiment. The drone has to
    navigate facing the blue net while moving from left to right.}
  \label{fig:res:paw_vs_pag_setup}
\end{figure}

We tested \method{} \gls{paw} capabilities against a naive \gls{pag}
planning method in a simple unobstructed point-to-point motion
scenario.  The path induces the drone to move across two textured
boxes which are not visible at the start. The setup is shown in
\figref{fig:res:paw_vs_pag_setup}.
The naive method does not account for any perception-related cost nor
constraint, thus producing much faster flight.  However, the system
struggles to track relevant features and quickly drifts accumulating
large errors.  \methodslam{} is able to factor these issues inherently
in the planning process, thus producing reliable trajectories that
favors safety and ultimately delivers better results.
Table~\ref{tab:paw_vs_pag_results} shows the tracking performance of
both methods when deploying \methodslam{}.
\begin{table}[t]
\centering
\caption{Comparison of PAG \& PAW planning results}
\setlength{\tabcolsep}{4pt}    
\renewcommand{\arraystretch}{0.98}
\label{tab:paw_vs_pag_results}
\begin{tabular}{l|l|cccccc}
\toprule
& Alg. & RMSE & Mean & Median & Std & Min & Max  \\
\midrule
\textbf{ATE}& PAG & 1.997 & 1.623 & 1.060 & 1.164 & 0.257 & 4.093 \\
\textbf{[m]}& PAW & \cellwin{0.674} & \cellwin{0.572} & \cellwin{0.408} & \cellwin{0.356} & \cellwin{0.177} & \cellwin{1.363} \\
\midrule
\textbf{RTE}& PAG & 2.900           & 1.807           & \cellwin{0.294} & 2.268           & \celldraw{0.000} & 5.147  \\
\textbf{[m]}& PAW & \cellwin{0.958} & \cellwin{0.705} & 0.486 & \cellwin{0.649} & \celldraw{0.000} & \cellwin{1.845} \\
\midrule
\textbf{RRE}& PAG & 2.69 & \cellwin{1.02} & \cellwin{0.35} & 2.49 & \celldraw{0.00} & 10.08 \\
\textbf{[deg]}& PAW & \cellwin{2.15} & 1.62 & 0.97 & \cellwin{1.41} & \celldraw{0.00} & \cellwin{5.89} \\
\bottomrule
\end{tabular}
\end{table}

\subsection{Full Pipeline Testing}

Finally, \method{} is tested in its entirety on multiple hardware
experiments, in the setup described in
Sec.~\ref{sec:experimental_results:setup}.  To verify the
effectiveness of the entire pipeline, the method is applied in a
\problemname{} motion relying solely on \methodslam{} to deliver robot
poses and the environmental map.

As stated in the previous sections, \methodslam{} provides improved
performance to ORB-SLAM3 in terms of estimation capabilities in the
demonstrated navigation tasks.  However, due to the nature of the
problem, the generated map and the estimated trajectories differ
significantly from the ground truth.  Nevertheless, the holistic
nature of the methodology is robust against drift and estimation
errors.  The method demonstrates to be able to reach the goal even
when subject to inaccurate estimation.  Moreover, it appropriately
matches the prescribed motion constraints, summarized in
Tab.~\ref{tab:experiment_constraints}.
\begin{table}[t]
\centering
\caption{Summary of Constraints}
\label{tab:experiment_constraints}
  \begin{tabular}{c|c|c|c|c|c}
    \toprule
    Constraint  &  $\angleaxisvector_{lim}$ & $\linearvelocity_{\max}$  & $\linearvelocity_{mean}$ & $\angularvelocity_{\max}$ &  $\pose_F \boxminus \posedesfinal$ \\
    \midrule
    \multirow{2}{*}{Value}       &  $\frac{\pi}{3}$      & 0.7           &  0.1           &  0.7            &  1.0  \\
                                  & \SI{}{rad}                  & $\SI{}{\metre\per\second}$ &  $\SI{}{\metre\per\second}$ &$\SI{}{rad\per\second}$ &  $\SI{}{\metre}$ \\
  \bottomrule
  \end{tabular}
\end{table}
Across $10$ experimental runs, the method consistently reaches the
goal region without collisions or failures, demonstrating the
robustness and effectiveness of the proposed approach in real-world
scenarios.

Moreover, Table~\ref{tab:minimal_metrics} provides an overview of the
performance of our method that consistently satisfies motion
constraint, up to a slight violation of the maximum angular velocity,
which can be partly attributed to aggressive replanning to maneuver
close to obstacles.  The final maps contains approximately $1000$
points, which are sufficient to represent minimally the textured parts
of the environment and maintain tracking.
\begin{table}[t]
\caption{Performance of \method{} in navigation tasks in an indoor environment.
Green indicates constraint satisfaction, yellow slight violation.}
\centering
\label{tab:minimal_metrics}
\setlength{\tabcolsep}{4pt}    
\renewcommand{\arraystretch}{0.98}
\begin{tabular}{l|c|c|c|c|c|c|c|c}
\toprule
\multirow{2}{*}{\textbf{R}}  
& dist.                     
& $\linearvelocity$         
& $\linearvelocity_{\max}$  
& $\angularvelocity$        
& $\angularvelocity_{\max}$  
& $\pose_F \boxminus \estPose_F$                 
& Map                       
& Time                      
\\
& $(m)$ & $(\frac{m}{s})$ & $(\frac{m}{s})$ & $(\frac{rad}{s})$ & $(\frac{rad}{s})$ & $(m)$ & \# & $(s)$  \\
\midrule
0 & 8.80  & 0.11  & \cellwin{0.54} & 0.06 & \cellwin{0.38}  & \cellwin{0.94} & 779  & 77   \\
1 & 9.82  & 0.14  & \cellwin{0.61} & 0.06 & \celldraw{0.89} & \cellwin{0.41} & 1067 & 72   \\
2 & 10.04 & 0.12  & \cellwin{0.66} & 0.06 & \cellwin{0.62}  & \cellwin{0.89} & 1107 & 86   \\
3 & 10.75 & 0.12  & \cellwin{0.46} & 0.05 & \cellwin{0.34}  & \cellwin{0.58} & 1061 & 92   \\
4 & 10.87 & 0.11  & \cellwin{0.65} & 0.07 & \celldraw{0.74} & \cellwin{0.84} & 1023 & 102  \\
5 & 11.15 & 0.07  & \cellwin{0.52} & 0.04 & \cellwin{0.21}  & \cellwin{0.66} & 1457 & 154  \\
6 & 10.74 & 0.10  & \cellwin{0.56} & 0.05 & \cellwin{0.29}  & \cellwin{0.66} & 1141 & 114  \\
7 & 12.86 & 0.13  & \cellwin{0.64} & 0.08 & \celldraw{0.78} & \cellwin{0.98} & 1239 & 112  \\
8 & 10.46 & 0.10  & \cellwin{0.63} & 0.04 & \cellwin{0.05}  & \cellwin{0.48} & 1456 & 110  \\
9 & 11.97 & 0.09  & \cellwin{0.46} & 0.04 & \cellwin{0.35}  & \cellwin{0.28} & 1185 & 139  \\
\bottomrule
\end{tabular}
\end{table}

\begin{figure}[t]
  \centering
    \begin{minipage}{0.8\columnwidth}
      \includegraphics[width=\linewidth]{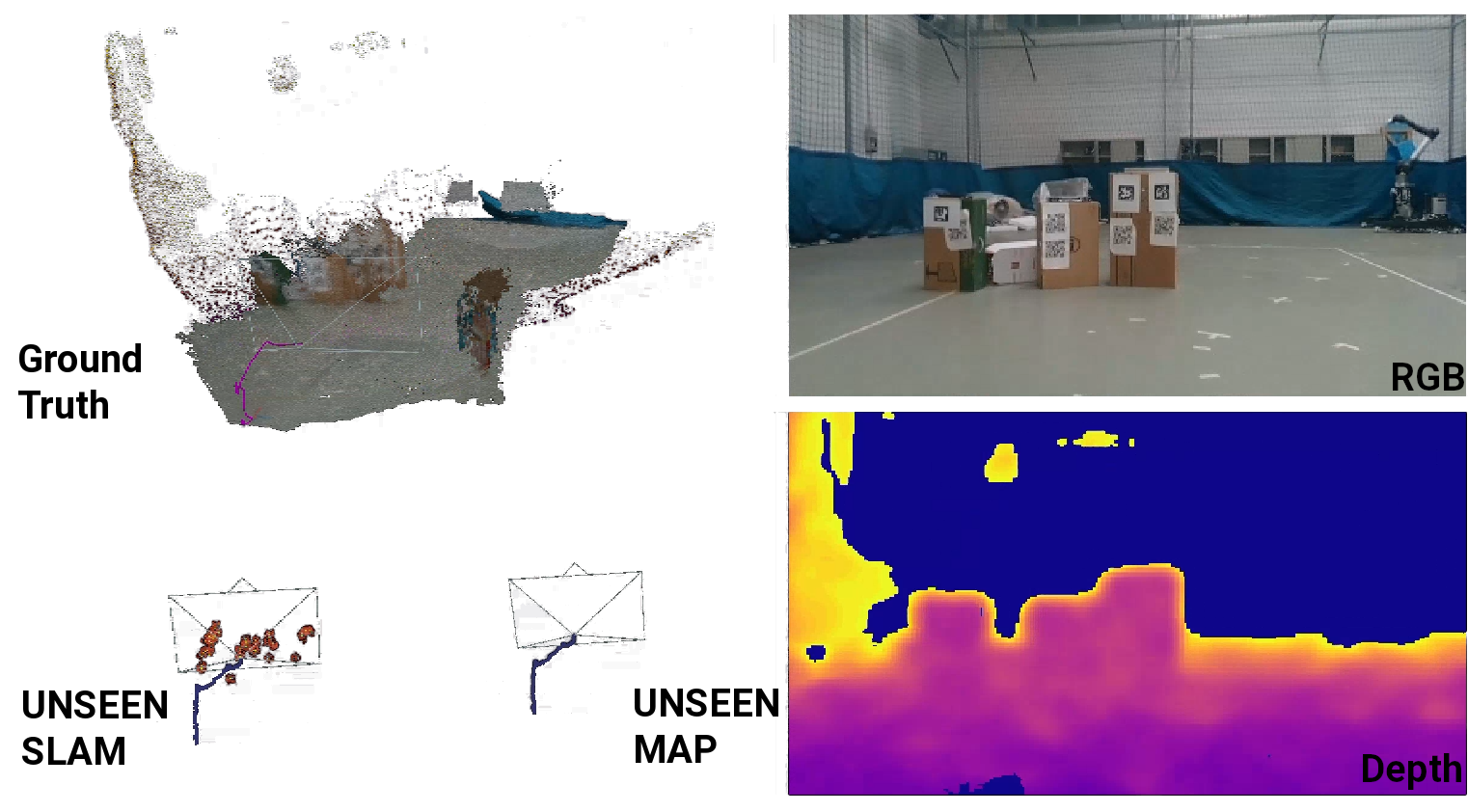}
    \end{minipage}
    \begin{minipage}{0.8\columnwidth}
      \includegraphics[width=\linewidth]{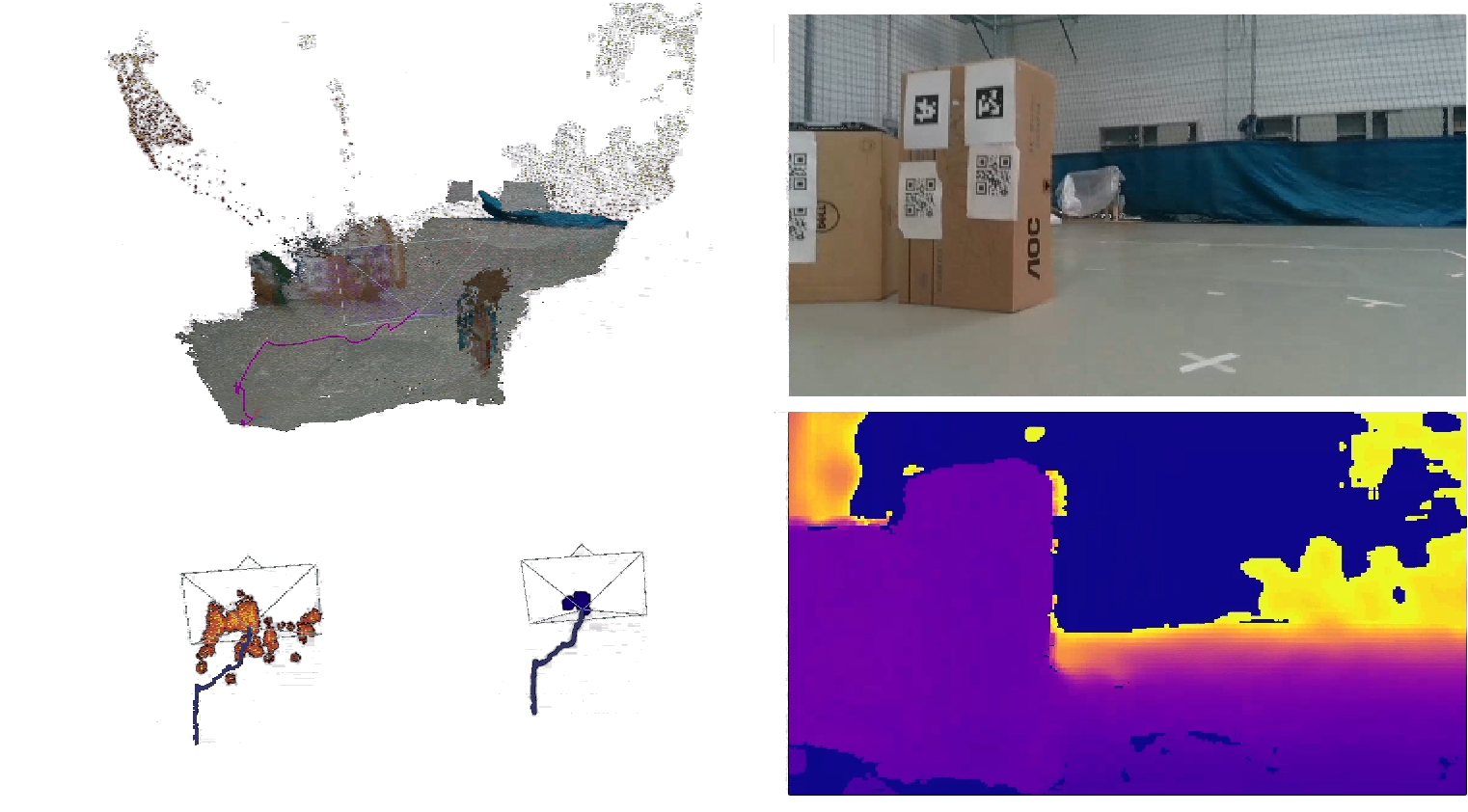}
    \end{minipage}
    \begin{minipage}{0.8\columnwidth}
      \includegraphics[width=\linewidth]{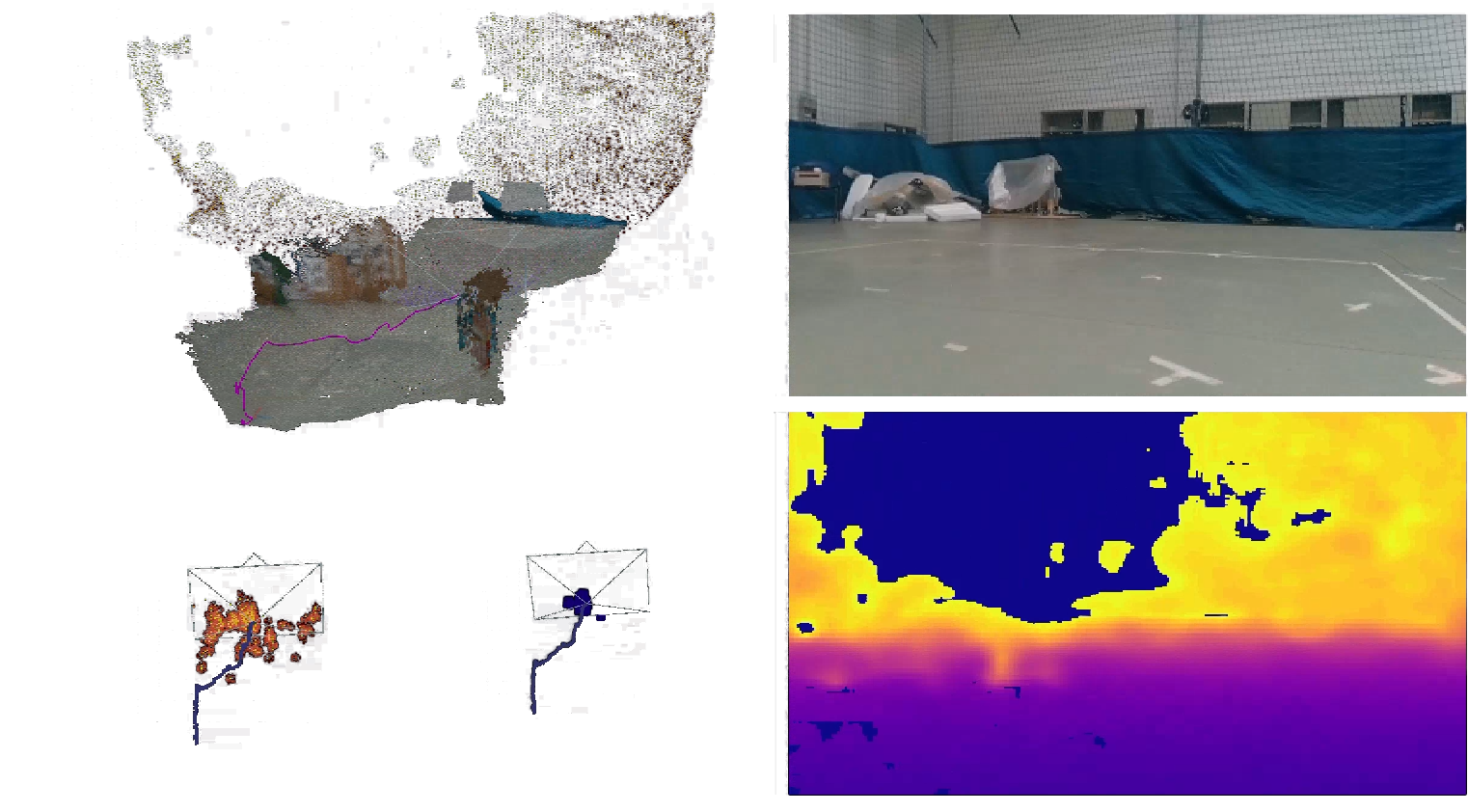}
    \end{minipage}
    \begin{minipage}{0.8\columnwidth}
      \includegraphics[width=\linewidth]{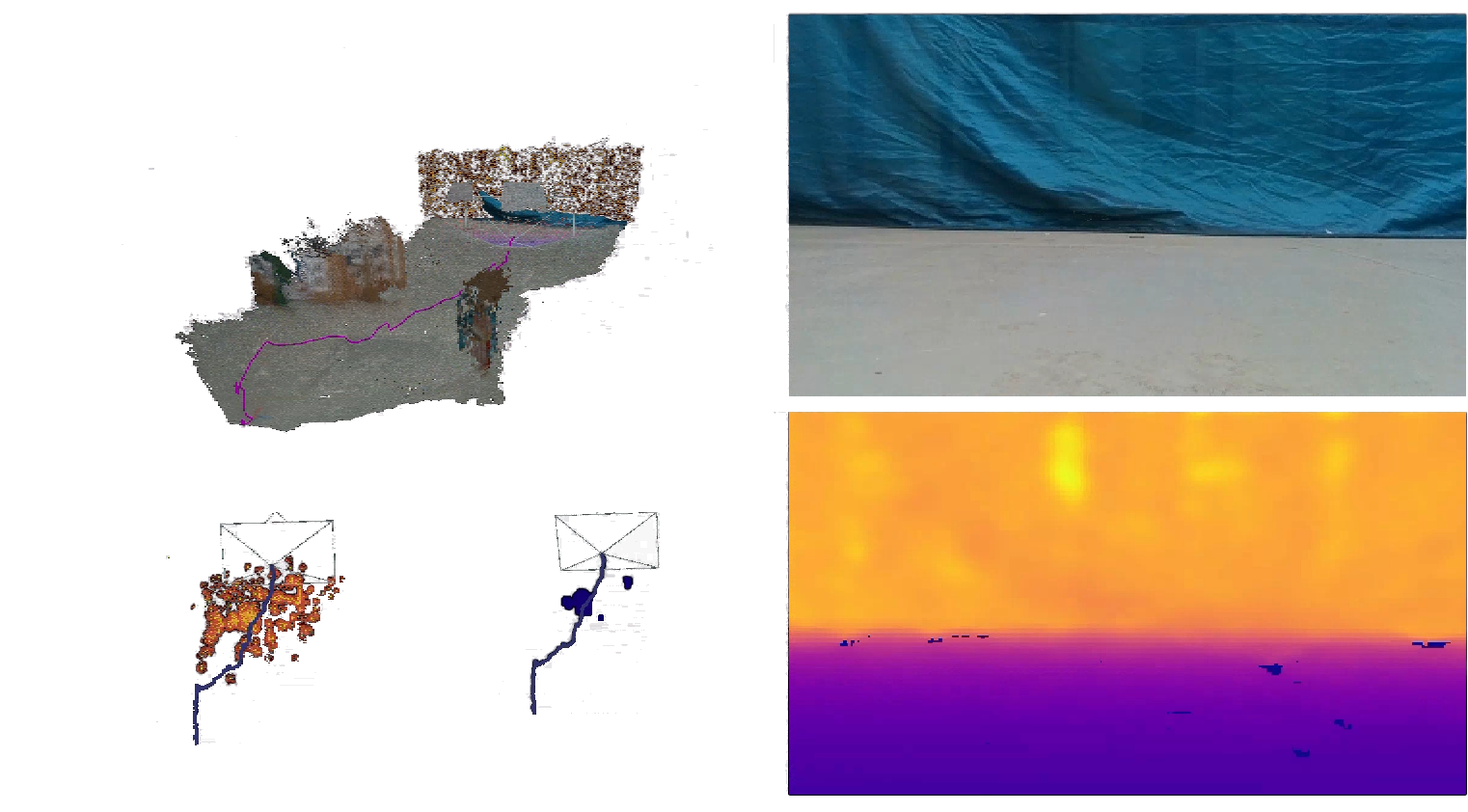}
    \end{minipage}
    \caption{
    Sequential snapshots of a flight performed with \method{}, from beginning (top) to end (bottom). 
    Ground Truth trajectory, reconstruction and current pointcloud (top left) obtained from the RGB-D images (right). \method{} operates online, performing \gls{slam} (bottom left) and occupancy mapping (bottom center). The robot navigates the passage while fixating on the textured obstacle. 
    Once the obstacle is overtaken, \method{} is able to reach the goal despite the lack of reliable features and the presence of a large white wall and a net, introducing errors even in the ground truth reconstruction.}
    \label{fig:res:experiment_hard:rerun}
\end{figure}

To further substantiate our analysis,
\figref{fig:res:experiment_hard:rerun} shows the snapshots of the
trajectory within the experimental setup.
The experiment showcases the uncertainty-aware nature of \method,
which approaches the well-textured obstacle in front of the camera, as
it allows to carve a portion of free space for the robot to traverse
(first row).  Later, the robot moves slightly sideways while
approaching the obstacle, to finally traverse the narrow gap and reach
the goal region (second row).  When the obstacle is overtaken (third
row), the environment is unexplored due to the previous occlusion and
the map becomes very noisy and uncertain due to the large amount of
unreliable texture present (fourth row). 

\begin{figure}[t]
    \centering
     \subfloat[]{\includegraphics[width=0.42\linewidth]{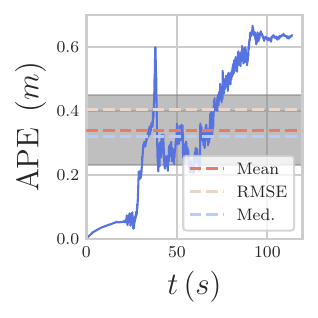}}   \quad 
     \subfloat[]{\includegraphics[width=0.42\linewidth]{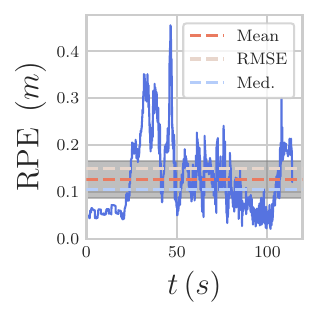}}  \\
     \subfloat[]{\includegraphics[width=0.42\linewidth]{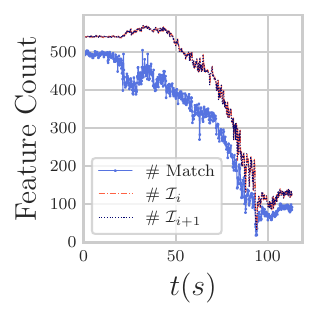}}      \quad
     \subfloat[]{\includegraphics[width=0.42\linewidth]{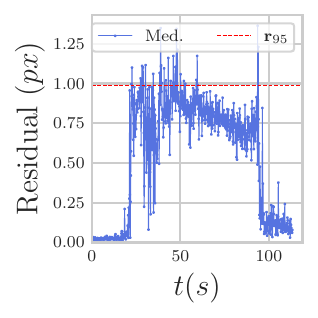}}        
     \caption{Performance of Visual SLAM in a Hard Narrow Passage
      experiment.  Development of \gls{APE} (a) and \gls{RPE} (b)
      overtime.  (c) and (d) show the number of features extracted,
      coupled with the median and 95th percentile value of residuals
      associated.}
    \label{fig:experiment_hard:metrics}
\end{figure}

\figref{fig:experiment_hard:metrics} showcases the correlation between
visual features and estimation error during flight.
The \gls{APE} shows a peak when the robot is avoiding the obstacle,
with a drop in feature matches and high variability of residuals.
During the approach and overtake ($28$~s mark), the \gls{APE} remains
stable showcasing the effectiveness of the local smoothing
employed. After overtaking it, the \gls{RPE} stabilizes, while the
\gls{APE} drifts significantly.  The consistent trend towards a lower
number of features extracted and matched correlates well with the
increased uncertainty of the resulting map and the progressive drift
in the trajectory.  Moreover, the increased number of large
covariances indicate that many landmarks are initialized and never
re-optimized due to inaccurate feature matching, highlighting once more how challenging the devised scenario can be when relying
solely on visual navigation.  Nevertheless, the system proves its
effectiveness and leverages the known uncertainty despite relying on
uncertainty-affected measurements.  As mentioned previously, despite the
map noise, the high uncertainty induces an increase of
the occupancy values within the voxels. 

In conclusion, the \method{} occupancy mapping approach shows its effectiveness in
clustering and capturing the uncertainty in the environment, ensuring
a traversable path for the robot through the narrow gap.

\section{Conclusions}

In this work, we introduced \method, a navigation pipeline that
tightly couples Localization, Mapping, and Planning to produce an
end-to-end, uncertainty-aware navigation system solely relying on a
front-facing RGB-D camera and sparse maps. 

We first developed
\methodslam{}, an uncertainty-driven SLAM module that provides both
poses and landmark estimates together with their covariances.  The
extracted uncertainty is directly incorporated into planning through
\methodmap{}, enabling consistent obstacle avoidance, occlusion
handling, and visibility queries.  These elements, combined with a
global path from \methodplan{}, feed into a receding-horizon strategy
on $\SEthree$ using perception-aware B-spline trajectory optimization
\methodtrajopt{}.  The resulting formulation introduces a
perception-aware cost that encourages trajectories maximizing
visibility of informative regions of the environment as it is
explored.

Each component was evaluated independently. \methodslam{} outperforms
standard ORB-SLAM3 in navigation tasks, delivering improvements in
\gls{ATE} up to 9\% and in \gls{RTE} up to 2\%.  \methodplan{} and
\methodtrajopt{} effectively work in simulation even with large maps
with up to 40'000 points when other state-of-the-art methods fail.
The proposed method computes feasible,
constraint-satisfying motions and leads to notably better state
estimation when compared to \gls{pag} and state-of-the-art \gls{paw}
methods. Finally, \method{} was deployed in 10 increasingly
challenging real-world unknown scenarios with limited reliable
textured areas.  It delivered 100\% success in reaching the goal
without collisions leveraging noisy measurements.  The method
demonstrated robust performance to satisfy prescribed motion
constraints, thus providing safe navigation to perform \problemname{}
mapping.

Future work will include an extension of the method to challenging
outdoor scenarios, thus considering environmental issues increasing
the measurement uncertainties, the inclusion of a wider class of
robotic platforms and the extension of the method to distributed
navigation and SLAM problems.

\bibliographystyle{IEEEtran}
\bibliography{IEEEabrv,src/ref}

\begin{thebibliography}{10}
\providecommand{\url}[1]{#1}
\csname url@samestyle\endcsname
\providecommand{\newblock}{\relax}
\providecommand{\bibinfo}[2]{#2}
\providecommand{\BIBentrySTDinterwordspacing}{\spaceskip=0pt\relax}
\providecommand{\BIBentryALTinterwordstretchfactor}{4}
\providecommand{\BIBentryALTinterwordspacing}{\spaceskip=\fontdimen2\font plus
\BIBentryALTinterwordstretchfactor\fontdimen3\font minus
  \fontdimen4\font\relax}
\providecommand{\BIBforeignlanguage}[2]{{%
\expandafter\ifx\csname l@#1\endcsname\relax
\typeout{** WARNING: IEEEtran.bst: No hyphenation pattern has been}%
\typeout{** loaded for the language `#1'. Using the pattern for}%
\typeout{** the default language instead.}%
\else
\language=\csname l@#1\endcsname
\fi
#2}}
\providecommand{\BIBdecl}{\relax}
\BIBdecl

\bibitem{tordesillas_panther_2022}
J.~Tordesillas and J.~P. How, ``{PANTHER}: {Perception}-{Aware} {Trajectory}
  {Planner} in {Dynamic} {Environments},'' \emph{IEEE Access}, vol.~10, pp.
  22\,662--22\,677, 2022.

\bibitem{bartolomei_semantic-aware_2021}
L.~Bartolomei, L.~Teixeira, and M.~Chli, ``Semantic-aware {Active} {Perception}
  for {UAVs} using {Deep} {Reinforcement} {Learning},'' in \emph{2021
  {IEEE}/{RSJ} {International} {Conference} on {Intelligent} {Robots} and
  {Systems} ({IROS})}, Sep. 2021, pp. 3101--3108.

\bibitem{zhou2021fuel}
B.~Zhou, Y.~Zhang, X.~Chen, and S.~Shen, ``Fuel: Fast uav exploration using
  incremental frontier structure and hierarchical planning,'' \emph{IEEE
  Robotics and Automation Letters}, vol.~6, no.~2, pp. 779--786, 2021.

\bibitem{kondo2024puma}
K.~Kondo, C.~T. Tewari, M.~B. Peterson, A.~Thomas, J.~Kinnari, A.~Tagliabue,
  and J.~P. How, ``Puma: Fully decentralized uncertainty-aware multiagent
  trajectory planner with real-time image segmentation-based frame alignment,''
  in \emph{2024 IEEE International Conference on Robotics and Automation
  (ICRA)}, 2024, pp. 13\,961--13\,967.

\bibitem{zhang_falcon_2024}
Y.~Zhang, X.~Chen, C.~Feng, B.~Zhou, and S.~Shen, ``Falcon: Fast autonomous
  aerial exploration using coverage path guidance,'' \emph{IEEE Transactions on
  Robotics}, vol.~41, pp. 1365--1385, 2025.

\bibitem{kondo_dynus_2025}
K.~Kondo, M.~Peterson, N.~Rober, J.~R. Viso, L.~Jia, J.~Chen, H.~Merton, and
  J.~P. How, ``Dynus: Uncertainty-aware trajectory planner in dynamic unknown
  environments,'' \emph{arXiv preprint arXiv:2504.16734}, 2025.

\bibitem{spechtLocalizationAware}
C.~Specht, T.~Faraci, V.~Kowalski~Martins, and R.~Lampariello,
  ``Localization-aware trajectory planning on se(3) for intravehicular
  robots,'' in \emph{2025 International Conference on Space Robotics, iSpaRo
  2025}, 2025.

\bibitem{chen_apace_2024}
X.~Chen, Y.~Zhang, B.~Zhou, and S.~Shen, ``Apace: Agile and perception-aware
  trajectory generation for quadrotor flights,'' in \emph{2024 IEEE
  International Conference on Robotics and Automation (ICRA)}, 2024, pp.
  17\,858--17\,864.

\bibitem{loquercio2021learning}
A.~Loquercio, E.~Kaufmann, R.~Ranftl, M.~Müller, V.~Koltun, and D.~Scaramuzza,
  ``Learning high-speed flight in the wild,'' \emph{Science Robotics}, vol.~6,
  no.~59, p. eabg5810, 2021.

\bibitem{kaufmann_champion_level_2023}
E.~Kaufmann, L.~Bauersfeld, A.~Loquercio, M.~Müller, V.~Koltun, and
  D.~Scaramuzza, ``Champion-level drone racing using deep reinforcement
  learning,'' \emph{Nature}, vol. 620, no. 7976, pp. 982--987, Aug. 2023.

\bibitem{davison2007monoslam}
A.~J. Davison, I.~D. Reid, N.~D. Molton, and O.~Stasse, ``Monoslam: Real-time
  single camera slam,'' \emph{IEEE Transactions on Pattern Analysis and Machine
  Intelligence}, vol.~29, no.~6, pp. 1052--1067, 2007.

\bibitem{bloesch2017rovio}
M.~Bloesch, M.~Burri, S.~Omari, M.~Hutter, and R.~Siegwart, ``Iterated extended
  kalman filter based visual-inertial odometry using direct photometric
  feedback,'' \emph{The International Journal of Robotics Research}, vol.~36,
  no.~10, pp. 1053--1072, 2017.

\bibitem{Geneva2020openvins}
P.~Geneva, K.~Eckenhoff, W.~Lee, Y.~Yang, and G.~Huang, ``Openvins: A research
  platform for visual-inertial estimation,'' in \emph{2020 IEEE International
  Conference on Robotics and Automation (ICRA)}, 2020, pp. 4666--4672.

\bibitem{mourikis2007multi}
A.~I. Mourikis and S.~I. Roumeliotis, ``A multi-state constraint kalman filter
  for vision-aided inertial navigation,'' in \emph{Proceedings 2007 IEEE
  International Conference on Robotics and Automation}, 2007, pp. 3565--3572.

\bibitem{rosinol2020kimera}
A.~Rosinol, M.~Abate, Y.~Chang, and L.~Carlone, ``Kimera: an open-source
  library for real-time metric-semantic localization and mapping,'' in
  \emph{2020 IEEE International Conference on Robotics and Automation (ICRA)},
  2020, pp. 1689--1696.

\bibitem{okvis2}
S.~Leutenegger, ``Okvis2: Realtime scalable visual-inertial slam with loop
  closure,'' 2022.

\bibitem{qin2019b}
T.~Qin, S.~Cao, J.~Pan, and S.~Shen, ``A general optimisation-based framework
  for global pose estimation with multiple sensors,'' \emph{IET Cyber-Systems
  and Robotics}, vol.~7, no.~1, p. e70023, 2025.

\bibitem{grisetti2011g2o}
R.~Kümmerle, G.~Grisetti, H.~Strasdat, K.~Konolige, and W.~Burgard, ``G2o: A
  general framework for graph optimization,'' in \emph{2011 IEEE International
  Conference on Robotics and Automation}, 2011, pp. 3607--3613.

\bibitem{isam2}
M.~Kaess, H.~Johannsson, R.~Roberts, V.~Ila, J.~Leonard, and F.~Dellaert,
  ``isam2: Incremental smoothing and mapping with fluid relinearization and
  incremental variable reordering,'' in \emph{2011 IEEE International
  Conference on Robotics and Automation}, 2011, pp. 3281--3288.

\bibitem{dellaert2017factor}
F.~Dellaert and M.~Kaess, ``Factor graphs for robot perception,''
  \emph{Foundations and Trends® in Robotics}, vol.~6, no. 1-2, pp. 1--139,
  2017.

\bibitem{mur2015orb}
R.~Mur-Artal, J.~M.~M. Montiel, and J.~D. Tardós, ``Orb-slam: A versatile and
  accurate monocular slam system,'' \emph{IEEE Transactions on Robotics},
  vol.~31, no.~5, pp. 1147--1163, 2015.

\bibitem{mur2017orb}
R.~Mur-Artal and J.~D. Tard\'os, ``{ORB-SLAM2}: an open-source {SLAM} system
  for monocular, stereo and {RGB-D} cameras,'' \emph{IEEE Transactions on
  Robotics}, vol.~33, no.~5, pp. 1255--1262, 2017.

\bibitem{orb_slam3}
C.~Campos, R.~Elvira, J.~J.~G. Rodríguez, J.~M. M.~Montiel, and J.~D.~Tardós,
  ``Orb-slam3: An accurate open-source library for visual, visual–inertial,
  and multimap slam,'' \emph{IEEE Transactions on Robotics}, 2021.

\bibitem{newcombe2011dtam}
R.~A. Newcombe, S.~J. Lovegrove, and A.~J. Davison, ``Dtam: Dense tracking and
  mapping in real-time,'' in \emph{2011 International Conference on Computer
  Vision}, 2011, pp. 2320--2327.

\bibitem{engel2015large}
J.~Engel, J.~St{\"u}ckler, and D.~Cremers, ``Large-scale direct slam with
  stereo cameras,'' in \emph{2015 IEEE/RSJ international conference on
  intelligent robots and systems (IROS)}.\hskip 1em plus 0.5em minus
  0.4em\relax IEEE, 2015, pp. 1935--1942.

\bibitem{newcombe2011kinectfusion}
R.~A. Newcombe, S.~Izadi, O.~Hilliges, D.~Molyneaux, D.~Kim, A.~J. Davison,
  P.~Kohi, J.~Shotton, S.~Hodges, and A.~Fitzgibbon, ``Kinectfusion: Real-time
  dense surface mapping and tracking,'' in \emph{2011 10th IEEE International
  Symposium on Mixed and Augmented Reality}, 2011, pp. 127--136.

\bibitem{whelan2016elasticfusion}
T.~Whelan, R.~F. Salas-Moreno, B.~Glocker, A.~J. Davison, and S.~Leutenegger,
  ``Elasticfusion: Real-time dense slam and light source estimation,''
  \emph{The International Journal of Robotics Research}, vol.~35, no.~14, pp.
  1697--1716, 2016.

\bibitem{han2019fiesta}
L.~Han, F.~Gao, B.~Zhou, and S.~Shen, ``Fiesta: Fast incremental euclidean
  distance fields for online motion planning of aerial robots,'' in \emph{2019
  IEEE/RSJ International Conference on Intelligent Robots and Systems (IROS)},
  2019, pp. 4423--4430.

\bibitem{hornung2013Octomap}
A.~Hornung, K.~M. Wurm, M.~Bennewitz, C.~Stachniss, and W.~Burgard, ``Octomap:
  an efficient probabilistic 3d mapping framework based on octrees,''
  \emph{Autonomous Robots}, vol.~34, no.~3, pp. 189--206, Apr 2013.

\bibitem{papatheodorou2025efficientsubmapbasedautonomousmav}
S.~Papatheodorou, S.~Boche, S.~B. Laina, and S.~Leutenegger, ``Efficient
  submap-based autonomous mav exploration using visual-inertial slam
  configurable for lidars or depth cameras,'' in \emph{2025 IEEE International
  Conference on Robotics and Automation (ICRA)}, 2025, pp. 7284--7290.

\bibitem{teed2021droid}
Z.~Teed and J.~Deng, ``Droid-slam: Deep visual slam for monocular, stereo, and
  rgb-d cameras,'' \emph{Advances in neural information processing systems},
  vol.~34, pp. 16\,558--16\,569, 2021.

\bibitem{qiu2025macvo}
Y.~Qiu, Y.~Chen, Z.~Zhang, W.~Wang, and S.~Scherer, ``Mac-vo: Metrics-aware
  covariance for learning-based stereo visual odometry mac-vo.github.io,'' in
  \emph{2025 IEEE International Conference on Robotics and Automation (ICRA)},
  2025, pp. 3803--3814.

\bibitem{duong_autonomous_2021}
T.~Duong, M.~Yip, and N.~Atanasov, ``Autonomous navigation in unknown
  environments with sparse bayesian kernel-based occupancy mapping,''
  \emph{IEEE Transactions on Robotics}, vol.~38, no.~6, pp. 3694--3712, 2022.

\bibitem{pairet2022safenav}
{\`E}.~Pairet, J.~D. Hern\'andez, M.~Carreras, Y.~Petillot, and M.~Lahijanian,
  ``Online mapping and motion planning under uncertainty for safe navigation in
  unknown environments,'' \emph{IEEE Transactions on Automation Science and
  Engineering}, vol.~19, no.~4, pp. 3356--3378, 2022.

\bibitem{oleynikova2017voxblox}
H.~Oleynikova, Z.~Taylor, M.~Fehr, R.~Siegwart, and J.~I. Nieto, ``Voxblox:
  Incremental 3d euclidean signed distance fields for on-board {MAV}
  planning,'' in \emph{2017 {IEEE/RSJ} International Conference on Intelligent
  Robots and Systems, {IROS} 2017, Vancouver, BC, Canada, September 24-28,
  2017}.\hskip 1em plus 0.5em minus 0.4em\relax {IEEE}, 2017, pp. 1366--1373.

\bibitem{vespa2018efficient}
E.~Vespa, N.~Nikolov, M.~Grimm, L.~Nardi, P.~H.~J. Kelly, and S.~Leutenegger,
  ``Efficient octree-based volumetric slam supporting signed-distance and
  occupancy mapping,'' \emph{IEEE Robotics and Automation Letters}, vol.~3,
  no.~2, pp. 1144--1151, 2018.

\bibitem{duberg2020ufomap}
D.~Duberg and P.~Jensfelt, ``Ufomap: An efficient probabilistic 3d mapping
  framework that embraces the unknown,'' \emph{IEEE Robotics and Automation
  Letters}, vol.~5, no.~4, pp. 6411--6418, 2020.

\bibitem{reijgwart2023wavemap}
V.~Reijgwart, C.~Cadena, R.~Siegwart, and L.~Ott, ``{Efficient volumetric
  mapping of multi-scale environments using wavelet-based compression},'' in
  \emph{Proceedings of Robotics: Science and Systems}, Daegu, Republic of
  Korea, July 2023.

\bibitem{fastautonomousflight2018}
K.~Mohta, M.~Watterson, Y.~Mulgaonkar, S.~Liu, C.~Qu, A.~Makineni, K.~Saulnier,
  K.~Sun, A.~Zhu, J.~Delmerico, K.~Karydis, N.~Atanasov, G.~Loianno,
  D.~Scaramuzza, K.~Daniilidis, C.~J. Taylor, and V.~Kumar, ``Fast, autonomous
  flight in gps-denied and cluttered environments,'' \emph{Journal of Field
  Robotics}, vol.~35, no.~1, pp. 101--120, 2018.

\bibitem{karaman2011sampling}
S.~Karaman and E.~Frazzoli, ``Sampling-based algorithms for optimal motion
  planning,'' \emph{The International Journal of Robotics Research}, vol.~30,
  no.~7, pp. 846--894, 2011.

\bibitem{zucker2013chomp}
M.~Zucker, N.~Ratliff, A.~D. Dragan, M.~Pivtoraiko, M.~Klingensmith, C.~M.
  Dellin, J.~A. Bagnell, and S.~S. Srinivasa, ``Chomp: Covariant hamiltonian
  optimization for motion planning,'' \emph{The International Journal of
  Robotics Research}, vol.~32, no. 9-10, pp. 1164--1193, 2013.

\bibitem{oleynikova2020open}
H.~Oleynikova, C.~Lanegger, Z.~Taylor, M.~Pantic, A.~Millane, R.~Siegwart, and
  J.~Nieto, ``An open-source system for vision-based micro-aerial vehicle
  mapping, planning, and flight in cluttered environments,'' \emph{Journal of
  Field Robotics}, vol.~37, no.~4, pp. 642--666, 2020.

\bibitem{chen2021navigablespace}
Z.~Chen and L.~Liu, ``Navigable space construction from sparse noisy point
  clouds,'' \emph{IEEE Robotics and Automation Letters}, vol.~6, no.~3, pp.
  4720--4727, 2021.

\bibitem{bircher2016receding}
A.~Bircher, M.~Kamel, K.~Alexis, H.~Oleynikova, and R.~Siegwart, ``Receding
  horizon "next-best-view" planner for 3d exploration,'' in \emph{2016 IEEE
  International Conference on Robotics and Automation (ICRA)}, 2016, pp.
  1462--1468.

\bibitem{Cieslewski2017}
T.~Cieslewski, E.~Kaufmann, and D.~Scaramuzza, ``Rapid exploration with
  multi-rotors: A frontier selection method for high speed flight,'' in
  \emph{2017 IEEE/RSJ International Conference on Intelligent Robots and
  Systems (IROS)}, 2017, pp. 2135--2142.

\bibitem{dharmadhikari2020motion}
M.~Dharmadhikari, T.~Dang, L.~Solanka, J.~Loje, H.~Nguyen, N.~Khedekar, and
  K.~Alexis, ``Motion primitives-based path planning for fast and agile
  exploration using aerial robots,'' in \emph{2020 IEEE International
  Conference on Robotics and Automation (ICRA)}, 2020, pp. 179--185.

\bibitem{yu2023echo}
J.~Yu, H.~Shen, J.~Xu, and T.~Zhang, ``Echo: An efficient heuristic viewpoint
  determination method on frontier-based autonomous exploration for
  quadrotors,'' \emph{IEEE Robotics and Automation Letters}, 2023.

\bibitem{yoder2016autonomous}
L.~Yoder and S.~Scherer, \emph{Autonomous Exploration for Infrastructure
  Modeling with a Micro Aerial Vehicle}.\hskip 1em plus 0.5em minus 0.4em\relax
  Cham: Springer International Publishing, 2016, pp. 427--440.

\bibitem{schmid2020efficient}
L.~Schmid, M.~Pantic, R.~Khanna, L.~Ott, R.~Siegwart, and J.~Nieto, ``An
  efficient sampling-based method for online informative path planning in
  unknown environments,'' \emph{IEEE Robotics and Automation Letters}, vol.~5,
  no.~2, pp. 1500--1507, 2020.

\bibitem{schmid_fast_2022}
L.~Schmid, C.~Ni, Y.~Zhong, R.~Siegwart, and O.~Andersson, ``Fast and
  {Compute}-{Efficient} {Sampling}-{Based} {Local} {Exploration} {Planning} via
  {Distribution} {Learning},'' \emph{IEEE Robotics and Automation Letters},
  vol.~7, no.~3, pp. 7810--7817, Jul. 2022.

\bibitem{zhang2022exploration}
Y.~Zhang, B.~Zhou, L.~Wang, and S.~Shen, ``Exploration with global consistency
  using real-time re-integration and active loop closure,'' in \emph{2022
  International Conference on Robotics and Automation (ICRA)}, 2022, pp.
  9682--9688.

\bibitem{song_learning_2023}
Y.~Song, K.~Shi, R.~Penicka, and D.~Scaramuzza, ``Learning perception-aware
  agile flight in cluttered environments,'' in \emph{2023 IEEE International
  Conference on Robotics and Automation (ICRA)}, 2023, pp. 1989--1995.

\bibitem{schmid_efficient_2020}
L.~Schmid, M.~Pantic, R.~Khanna, L.~Ott, R.~Siegwart, and J.~Nieto, ``An
  {Efficient} {Sampling}-{Based} {Method} for {Online} {Informative} {Path}
  {Planning} in {Unknown} {Environments},'' \emph{IEEE Robotics and Automation
  Letters}, vol.~5, no.~2, pp. 1500--1507, Apr. 2020.

\bibitem{moon2025iatigris}
B.~Moon, N.~Suvarna, A.~Jong, S.~Chatterjee, J.~Yuan, M.~Cao, and S.~Scherer,
  ``Ia-tigris: An incremental and adaptive sampling-based planner for online
  informative path planning,'' 2025.

\bibitem{palazzolo2018effective}
E.~Palazzolo and C.~Stachniss, ``Effective exploration for mavs based on the
  expected information gain,'' \emph{Drones}, 2018.

\bibitem{Papachristos2017icra}
C.~Papachristos, S.~Khattak, and K.~Alexis, ``Uncertainty-aware receding
  horizon exploration and mapping using aerial robots,'' in \emph{2017 IEEE
  International Conference on Robotics and Automation (ICRA)}, 2017.

\bibitem{papachristos_localization_2019}
C.~Papachristos, F.~Mascarich, S.~Khattak, T.~Dang, and K.~Alexis,
  ``Localization uncertainty-aware autonomous exploration and mapping with
  aerial robots using receding horizon path-planning,'' \emph{Autonomous
  Robots}, vol.~43, no.~8, pp. 2131--2161, Dec 2019.

\bibitem{zhang2018perception}
Z.~Zhang and D.~Scaramuzza, ``Perception-aware receding horizon navigation for
  mavs,'' in \emph{2018 IEEE International Conference on Robotics and
  Automation (ICRA)}.\hskip 1em plus 0.5em minus 0.4em\relax IEEE, 2018.

\bibitem{bartolomei_perception-aware_2020}
L.~Bartolomei, L.~Teixeira, and M.~Chli, ``Perception-aware {Path} {Planning}
  for {UAVs} using {Semantic} {Segmentation},'' in \emph{2020 {IEEE}/{RSJ}
  {International} {Conference} on {Intelligent} {Robots} and {Systems}
  ({IROS})}, Oct. 2020, pp. 5808--5815.

\bibitem{murali2019perception}
V.~Murali, I.~Spasojevic, W.~Guerra, and S.~Karaman, ``Perception-aware
  trajectory generation for aggressive quadrotor flight using differential
  flatness,'' in \emph{2019 American Control Conference (ACC)}, 2019, pp.
  3936--3943.

\bibitem{spasojevic_perception-aware_2020}
I.~Spasojevic, V.~Murali, and S.~Karaman, ``Perception-aware time optimal path
  parameterization for quadrotors,'' in \emph{2020 IEEE International
  Conference on Robotics and Automation (ICRA)}, 2020, pp. 3213--3219.

\bibitem{Boche_2025}
S.~Boche, J.~Jung, S.~B. Laina, and S.~Leutenegger, ``Okvis2-x: Open
  keyframe-based visual-inertial slam configurable with dense depth or lidar,
  and gnss,'' \emph{IEEE Transactions on Robotics}, vol.~41, p. 6064–6083,
  2025.

\bibitem{zhang2025falcon}
Y.~Zhang, X.~Chen, C.~Feng, B.~Zhou, and S.~Shen, ``Falcon: Fast autonomous
  aerial exploration using coverage path guidance,'' \emph{IEEE Transactions on
  Robotics}, vol.~41, pp. 1365--1385, 2025.

\bibitem{liu2025tikhonov}
X.~Liu and M.~Cao, ``High-order regularization dealing with ill-conditioned
  robot localization problems,'' \emph{IEEE Transactions on Robotics}, vol.~41,
  pp. 3539--3555, 2025.

\bibitem{forster2016manifold}
C.~Forster, L.~Carlone, F.~Dellaert, and D.~Scaramuzza, ``On-manifold
  preintegration for real-time visual-inertial odometry,'' \emph{IEEE
  Transactions on Robotics}, vol.~33, no.~1, pp. 1--21, 2016.

\bibitem{hartley2003multiple}
R.~Hartley and A.~Zisserman, \emph{Multiple View Geometry in Computer Vision},
  2nd~ed.\hskip 1em plus 0.5em minus 0.4em\relax Cambridge University Press,
  2004.

\bibitem{herlihy20120multiprocessor_programming}
M.~Herlihy and N.~Shavit, \emph{The Art of Multiprocessor Programming, Revised
  Reprint}, 1st~ed.\hskip 1em plus 0.5em minus 0.4em\relax San Francisco, CA,
  USA: Morgan Kaufmann Publishers Inc., 2012.

\bibitem{sucan2012the-open-motion-planning-library}
I.~A. Sucan, M.~Moll, and L.~E. Kavraki, ``The open motion planning library,''
  \emph{IEEE Robotics \& Automation Magazine}, vol.~19, no.~4, pp. 72--82,
  2012.

\bibitem{liu2017decomputil}
S.~Liu, M.~Watterson, K.~Mohta, K.~Sun, S.~Bhattacharya, C.~J. Taylor, and
  V.~Kumar, ``Planning dynamically feasible trajectories for quadrotors using
  safe flight corridors in 3-d complex environments,'' \emph{IEEE Robotics and
  Automation Letters}, vol.~2, no.~3, pp. 1688--1695, 2017.

\bibitem{kume2009fisher}
A.~Kume and T.~Sei, ``On the fisher--bingham distribution,'' \emph{Statistics
  and Computing}, vol.~19, no.~2, pp. 167--172, 2009.

\bibitem{brady1982robot}
M.~Brady, \emph{Robot motion: Planning and control}.\hskip 1em plus 0.5em minus
  0.4em\relax MIT press, 1982.

\bibitem{faraci2025reachability}
T.~Faraci and R.~Lampariello, ``Reachability-guaranteed optimal control for the
  interception of dynamic targets under uncertainty,'' in \emph{2025 European
  Control Conference (ECC)}, 2025, pp. 2824--2831.

\bibitem{sommer2020efficient}
C.~Sommer, V.~Usenko, D.~Schubert, N.~Demmel, and D.~Cremers, ``Efficient
  derivative computation for cumulative b-splines on lie groups,'' in
  \emph{Proceedings of the IEEE/CVF conference on computer vision and pattern
  recognition}, 2020, pp. 11\,148--11\,156.

\bibitem{rossmann2006lie}
W.~Rossmann, \emph{Lie Groups: An Introduction Through Linear Groups}.\hskip
  1em plus 0.5em minus 0.4em\relax Oxford University Press, 01 2002.

\end{thebibliography}

\vspace{11pt}

\end{document}